\documentclass{article}
\usepackage[OT1]{fontenc}
\usepackage[letterpaper,hmargin=1.5in,vmargin=1in]{geometry}
\usepackage{times}
\usepackage{natbib}
\setcitestyle{authoryear,round,citesep={;},aysep={,},yysep={;}}
\usepackage{amsmath,amssymb,bm}
\usepackage{xcolor}
\newcommand{\bestmean}[1]{\ensuremath{\mathbf{#1}}}
\newcommand{\tablesd}[1]{\text{\normalfont\footnotesize\color{black!65}\ensuremath{\pm #1}}}
\usepackage{booktabs}
\usepackage{array}
\usepackage{tabularx}
\usepackage{graphicx}
\usepackage{float}
\usepackage{wrapfig}
\newif\ifshowpaperfigures
\showpaperfiguresfalse
\usepackage{tikz}
\usetikzlibrary{arrows.meta,positioning,fit,calc}
\usepackage{hyperref}
\usepackage{url}

\definecolor{accent}{HTML}{2457A6}
\definecolor{muted}{HTML}{666666}
\definecolor{placeholder}{HTML}{A33A2B}
\definecolor{attrweight}{HTML}{2F6BFF}
\definecolor{attrsize}{HTML}{D97904}
\definecolor{attrfold}{HTML}{7656B7}
\definecolor{creditpos}{HTML}{21854A}
\definecolor{creditneg}{HTML}{C23B3B}
\definecolor{fallbackgray}{HTML}{73777F}

\hypersetup{
  pdfauthor={Gen Li, Pengyang Xie, Xiao Xiao, Chuanzhi Tong, Ziyang Ji, Zhigang Xu, Huiwen Bai, Zihao Xing, Yanxiang Zhou, Yifei Yuan, Xiaoqing Wang},
  pdftitle={AnchorCredit: Output-Anchored Credit Assignment for Structured Generation with Multiple Verifiable Rewards}
}

\newcommand{\method}{AnchorCredit}

\title{AnchorCredit: Output-Anchored Credit Assignment for Structured Generation with Multiple Verifiable Rewards}
\author{%
  Gen Li, Pengyang Xie\thanks{Corresponding author. \href{mailto:xiepengyang.xpy@alibaba-inc.com}{\texttt{xiepengyang.xpy@alibaba-inc.com}}\enspace\textperiodcentered\enspace\href{https://github.com/AlibabaResearch/anchorcredit}{\textcolor{accent}{Code on GitHub}}}, Xiao Xiao, Chuanzhi Tong, Ziyang Ji, Zhigang Xu \\
  Huiwen Bai, Zihao Xing, Yanxiang Zhou, Yifei Yuan, Xiaoqing Wang \\[0.4em]
  Alibaba Group
}
\date{}

\begin{document}

\maketitle

\begin{abstract}
In Reinforcement Learning with Verifiable Rewards (RLVR), Group Relative Policy Optimization (GRPO) and related methods typically compute group-relative advantages from response rewards and broadcast the same advantage to every generated token in a response to guide policy updates. However, we identify a systematic problem of cross-component credit misassignment in tasks with multiple verifiable results, such as joint multi-attribute prediction: low rewards for some components can penalize a well-performing component and its associated reasoning. To address this problem, we propose \textbf{\method{}}. The method uses the policy model's attention from output anchors to preceding reasoning sentences to estimate sentence--component associations and assign component-specific advantages to the relevant reasoning, without additional relevance annotations or a separate relevance discriminator. To support the study of credit assignment with multiple verifiable rewards, we construct ProductAttr, a benchmark of 7,200 real product image--text samples. Experiments show that \method{} outperforms existing baselines on this benchmark. Experiments on the public Nutrition5k dataset further support the method's effectiveness in this setting.
\end{abstract}

\section{Introduction}

Reinforcement Learning with Verifiable Rewards (RLVR) has substantially improved the reasoning capabilities of large language models on automatically verifiable tasks such as mathematics and coding~\citep{shao2024deepseekmath,deepseekai2025r1}. Existing methods typically evaluate an entire response with a scalar reward. However, many tasks require a model to generate reasoning followed by a structured output containing multiple verifiable output components in a single response. These components may represent product attributes or dialogue slots~\citep{yang2021mave,rastogi2020towards}. For example, a product understanding task may require jointly predicting several attributes, such as weight, packaged dimensions, and foldability, from product images and text. We study this setting, in which each output component can be verified separately.

When Group Relative Policy Optimization (GRPO) is applied to these tasks, the rewards for the components of each response are weighted and summed to obtain a single response-level reward. These total rewards are standardized across responses sampled for the same prompt to obtain response-level advantages~\citep{shao2024deepseekmath}. The same response-level advantage is then used to weight the policy-loss contribution of every generated token in that response. However, a shared response-level advantage can penalize a component whose own advantage is positive. In Figure~\ref{fig:leakage-overview}(a), errors in weight and dimensions result in a negative response-level advantage, causing GRPO to penalize the correct foldability prediction and its associated reasoning despite a positive foldability-specific advantage. We refer to this mismatch as \textbf{cross-component credit misassignment}.

\ifshowpaperfigures
\else
\begin{figure}[t]
\centering
\includegraphics[width=\textwidth]{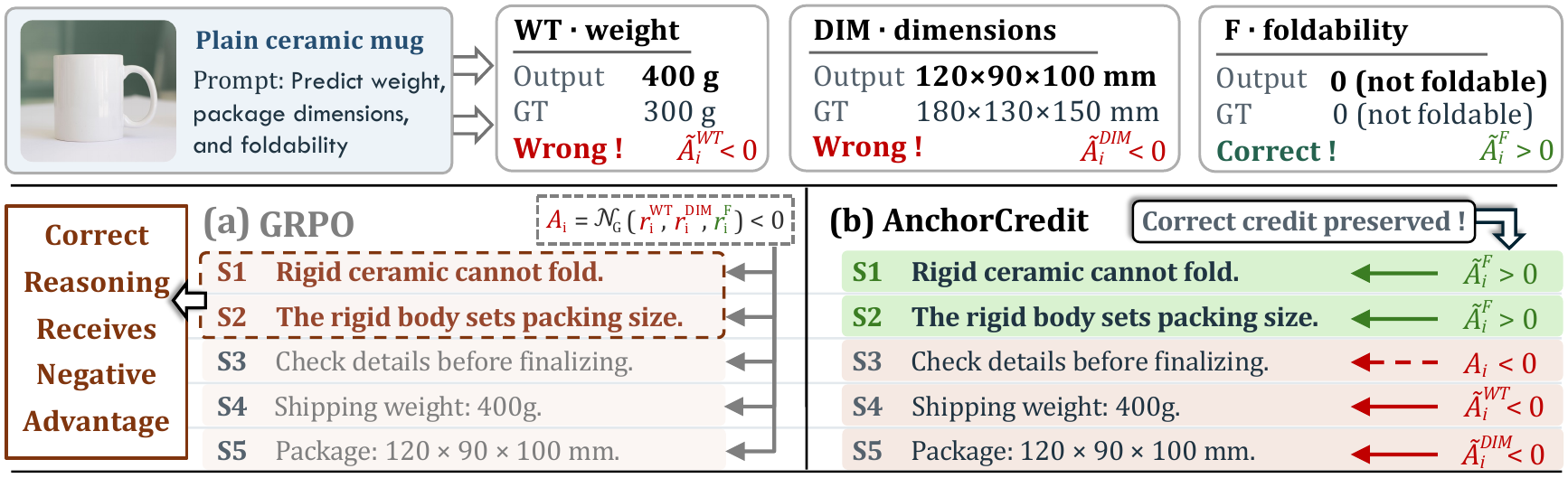}
\caption{Cross-component credit misassignment and \method{}. (a) A shared negative advantage penalizes correct foldability reasoning. (b) \method{} assigns component-specific credit to associated sentences. Signs are shown before final whitening.}
\label{fig:leakage-overview}
\end{figure}
\fi

Although each output component can be scored separately, these scores do not specify which preceding reasoning sentences should receive its learning signal. Recent work has explored alternatives to GRPO for multi-reward optimization. GDPO~\citep{liu2026gdpo} normalizes rewards separately before aggregation, while GD$^2$PO~\citep{liu2026gd2po} filters conflicting rollouts. However, both retain response-level advantages and leave this attribution problem unresolved. RTT~\citep{xu2026rtt} maps rubric feedback to individual tokens through a relevance discriminator, but training this discriminator requires additional token-level relevance annotations and a separate training stage.

To address these limitations, we propose \textbf{\method{}}, a plug-in module for component-specific credit assignment. Because each generated output value corresponds to a known component, we use its tokens as anchors. The policy model's attention from these anchors to preceding reasoning sentences provides an estimate of how strongly each sentence is associated with the corresponding component. These associations guide the assignment of component-specific advantages to the relevant sentences (Figure~\ref{fig:leakage-overview}(b)), without additional relevance annotations or a separate relevance discriminator. The resulting token-level credit can be incorporated into the policy update of an existing training algorithm.

We introduce ProductAttr, a benchmark of 7,200 product image--text samples for joint prediction of continuous and categorical attributes. Adding \method{} to GRPO improves mean performance on all three target attributes in both ProductAttr and Nutrition5k~\citep{thames2021nutrition5k}. On ProductAttr, the improvements in weight RMAE, dimension RMAE, and foldability F1 are 5.74, 2.80, and 1.76 percentage points, respectively. These gains extend to dialogue state tracking on Schema-Guided Dialogue (SGD; \citealp{rastogi2020towards}), where \method{} improves Joint Goal Accuracy over GRPO by 6.00 percentage points. Gains with DAPO and GDPO show that the benefits extend beyond GRPO, while improvements on larger dense and MoE models further support applicability across model configurations.

In summary, we identify cross-component credit misassignment, where a shared response-level advantage can oppose a component-specific learning signal. To address this mismatch, we propose \method{}, a plug-in module that uses attention from output anchors to assign component-specific credit to reasoning sentences. We also construct ProductAttr with fixed splits and component-wise verifiers, and evaluate the module on multi-attribute prediction and dialogue state tracking.

\section{Related Work}

\paragraph{Reinforcement learning with verifiable rewards.}

RLVR uses automatically verifiable rewards for tasks such as mathematics and coding~\citep{shao2024deepseekmath,deepseekai2025r1}. GRPO computes group-relative advantages without a separately trained value model~\citep{shao2024deepseekmath}; DAPO improves training through asymmetric clipping, dynamic sampling, and a token-level loss~\citep{yu2025dapo}. Both construct advantages from response-level rewards. \method{} instead assigns component-specific credit to relevant positions while retaining the underlying algorithm's sampling and policy-update mechanisms.

\paragraph{Fine-grained credit assignment.}

Process reward models learn from human or automatically constructed step labels~\citep{lightman2024verify,wang2024mathshepherd}. Other approaches estimate local credit through progress toward success, prefix rollouts, or segment-level advantages~\citep{setlur2025rewarding,kazemnejad2025vineppo,guo2025spo}. Attention-based approaches redistribute response rewards through reward-model attention~\citep{chan2024dense}, amplify advantages at selected positions~\citep{li2025attention}, or weight token losses by attention flow toward the answer, as in FlowTracer~\citep{dong2026flowtracer}. For multiple feedback signals, RTT maps rubric feedback to token rewards using a separately trained relevance discriminator~\citep{xu2026rtt}; SRaR uses an LLM judge to associate rubric items with reasoning steps~\citep{xie2026stepwise}; BAE assigns objective-specific advantages to predefined text blocks~\citep{pavlenko2026blockwise}. \method{} instead uses generated output values and the policy model's attention to estimate sentence--component associations and assign component-specific credit, without predefined reasoning blocks, relevance annotations, or a separate relevance model.

\paragraph{Multidimensional rewards and structured prediction.}

In multi-reward language-model training, GDPO standardizes each reward within the group before aggregating the advantages~\citep{liu2026gdpo}. GD$^2$PO additionally filters and reweights rollouts and queries based on conflicting reward-advantage signs~\citep{liu2026gd2po}. Both retain response-level signals; \method{} complements reward aggregation by assigning component-specific feedback to relevant content positions. MAVE, visual physical-attribute datasets, and Nutrition5k support product attribute extraction, physical-attribute prediction, and nutrition estimation~\citep{yang2021mave,zhai2024physical,thames2021nutrition5k}. Their original settings emphasize supervised prediction; we study how reinforcement learning assigns multiple attribute feedback signals within a generated response.

\section{Problem Formulation}\label{sec:problem-formulation}

We consider tasks in which a model generates reasoning followed by a structured output with $K$ separately verifiable components. For an input $x$, the policy $\pi_\theta$ generates a group of responses. Let $r_i^k$ denote the reward assigned by the verifier to component $k$ of response $i$.

GRPO-style methods aggregate component rewards into a scalar reward and standardize it within the group to obtain a response-level advantage. This advantage measures overall performance but does not distinguish the relative performance of each component. Following GDPO~\citep{liu2026gdpo}, we standardize each component's rewards separately within the group. We use the resulting component-specific advantage to measure how a response performs on that component relative to other responses in the group:
\begin{equation}
 A_i=\mathcal N_G\!\bigl(\textstyle\sum_{k=1}^{K}\omega_k r_i^k\bigr),\qquad
 A_i^k=\mathcal N_G\!\bigl(r_i^k\bigr),\quad k=1,\ldots,K.
 \label{eq:group-advantages}
\end{equation}
Here, $\omega_k$ denotes the predefined task weight for component $k$, and $\mathcal N_G$ denotes standardization across responses sampled for the same input. $A_i$ measures relative overall performance.

GRPO applies the same response-level advantage $A_i$ to every generated token~\citep{shao2024deepseekmath}. When $A_iA_i^k<0$, this shared signal opposes the component-specific advantage, formalizing the cross-component credit misassignment illustrated in Figure~\ref{fig:leakage-overview}(a).

To assign component-specific credit to reasoning sentences, we need to identify which sentences are relevant to each component. For component $k$, we use the tokens of its generated value as \textbf{output anchors} and denote their positions in response $i$ by $S_i^k$. However, which preceding reasoning sentences are relevant to each component is not specified in advance.

We therefore study how to use output anchors to associate reasoning sentences with verifiable output components and assign credit to those sentences based on the corresponding component-specific advantages. For sentences associated with multiple components, we also need to determine how to combine their advantages.

\section{AnchorCredit: Output-Anchored Credit Assignment}

\method{} constructs token-level credit in three steps: computing component-specific advantages, estimating sentence--component associations using output anchors, and applying association gating and sign-aware combination to obtain sentence credit (Figure~\ref{fig:method-overview}). Anchor tokens receive the advantage of their corresponding component directly, while the tokens in each reasoning sentence receive that sentence's credit.

\ifshowpaperfigures
\else
\begin{figure}[t] \centering \includegraphics[width=\textwidth]{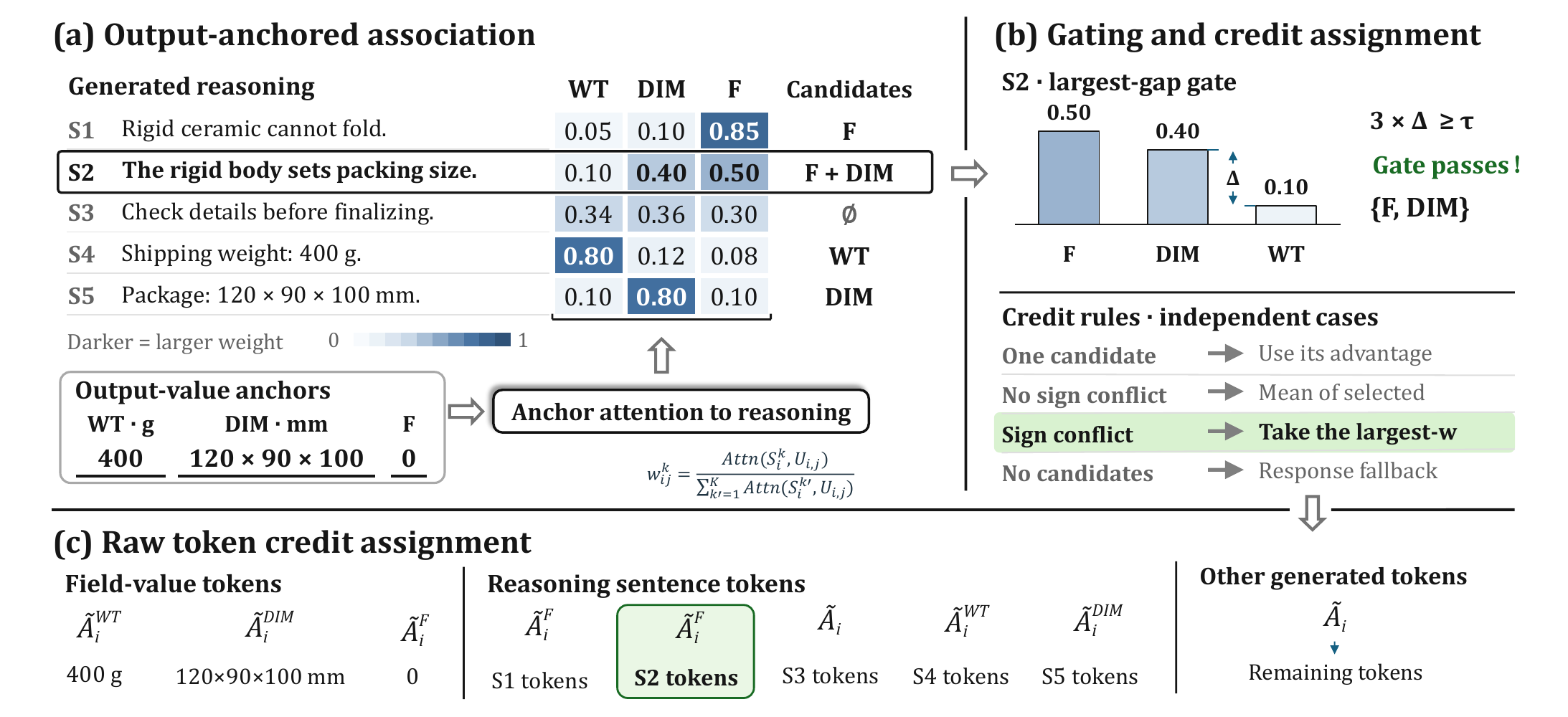} \caption{Output-anchored credit assignment. (a) Attention from output-value anchors estimates sentence--component associations. (b) Largest-gap gating selects candidate components; the highlighted example uses the conflicting-sign rule. (c) Component-specific and fallback advantages are assembled into raw token credit before global whitening. Association weights are illustrative.} \label{fig:method-overview} \end{figure}
\fi

\subsection{From Component-Specific Rewards to Component-Specific Advantages}

\method{} starts from the component-specific group-relative advantages $A_i^k$ defined in Equation~\eqref{eq:group-advantages}, computed by standardizing the rewards for each component separately across responses to the same input.

Unlike GDPO~\citep{liu2026gdpo}, which aggregates component advantages immediately and whitens only their sum, \method{} routes separate component advantages, compares their signs, and falls back to $A_i$. Because components differ in batch variance after group standardization (e.g., binary foldability yields more zero-variance groups), we whiten each component column and $A_i$ before routing:
\begin{equation}
 \widetilde A_i^k=\operatorname{Whiten}(A^k)_i,\qquad
 \widetilde A_i=\operatorname{Whiten}(A)_i.
 \label{eq:pre-routing-whitening}
\end{equation}
Centering may change advantage signs; all subsequent sign comparisons and routing use the calibrated $\widetilde A_i^k$ specifying the sign and magnitude of credit for component $k$. We next estimate which reasoning sentences receive credit from each component; unrouted positions retain $\widetilde A_i$ before final whitening.

\subsection{Attention-Based Sentence--Component Association Estimation}

We use the anchor-token positions $S_i^k$ defined in Section~\ref{sec:problem-formulation} to estimate which reasoning sentences are associated with component $k$.

Because output anchors follow the reasoning text, we use their attention to preceding tokens to estimate sentence--component associations. We divide the reasoning text of response $i$ into nonempty sentences using boundaries such as sentence-final punctuation and blank lines. We denote sentence $j$ by $U_{i,j}$ and estimate associations at the sentence level. Table~\ref{tab:routing-granularity} in Appendix~\ref{sec:online-routing-controls} compares the end-to-end results of sentence-level and token-level routing; the appendix gives the sentence-splitting rules.

Let $\alpha_{st}^{\ell h}$ denote the attention from anchor token $s$ to preceding reasoning token $t$ in head $h$ of layer $\ell$. Let $\mathcal L$ be the full-attention layers from which explicit attention weights are extracted, and let $H$ be the number of heads in each such layer. Here, $U_{i,j}$ also denotes the token positions of sentence $j$. We average equally over these layers, heads, anchor tokens, and sentence tokens:
\begin{equation}
 \operatorname{Attn}(S_i^k,U_{i,j})
 =\frac{1}{|\mathcal L|H|S_i^k||U_{i,j}|}
 \sum_{\ell\in\mathcal L}\sum_{h=1}^{H}
 \sum_{s\in S_i^k}\sum_{t\in U_{i,j}}\alpha_{st}^{\ell h}.
 \label{eq:attention-pooling}
\end{equation}
For each reasoning sentence, we normalize these attention scores across components to obtain the association weights:
\begin{equation}
 w_{ij}^k
 =\frac{\operatorname{Attn}(S_i^k,U_{i,j})}
 {\sum_{k'=1}^{K}\operatorname{Attn}(S_i^{k'},U_{i,j})},
 \qquad k=1,\ldots,K.
 \label{eq:association}
\end{equation}
The weight $w_{ij}^k$ estimates the relative association between sentence $j$ and component $k$. The schema determines which component an anchor belongs to. We use attention as an association estimate for credit routing~\citep{jain2019attention,wiegreffe2019attention}. Nearly uniform weights do not distinguish a preferred component, while multiple selected components may have advantages with opposing signs. We therefore use a gate to select candidate components before combining their advantages.

The separation between anchor-token credit and reasoning routing is specified in the credit-assembly step below.

\subsection{Candidate-Component Gating and Signed Credit Assignment}

The preceding steps provide an advantage $\widetilde A_i^k$ for each component and association weights $w_{ij}^k$ for each reasoning sentence. \method{} uses these weights to select candidate components and combines their advantages according to their signs. The resulting sentence credit is assigned to every token in that sentence, while the remaining tokens are handled in the credit-assembly step.

\paragraph{Candidate selection using the normalized largest gap.}
A top-1 rule selects one component even when the association weights are nearly uniform. Instead, we select candidates using the largest gap between adjacent sorted weights. Let $w_{ij}^{(1)}\ge\cdots\ge w_{ij}^{(K)}$ denote the weights in descending order, and define
\begin{equation}
 \begin{aligned}
 g_m&=w_{ij}^{(m)}-w_{ij}^{(m+1)}, &&m=1,\ldots,K-1,\\
 m^*&=\min\{m:g_m=\max_{1\le q<K}g_q\},
 &\widetilde g_{ij}&=K g_{m^*}.
 \end{aligned}
 \label{eq:normalized-gap}
\end{equation}
Only when $\widetilde g_{ij}\ge\tau$ does \method{} retain components before the largest gap whose association weights strictly exceed the uniform baseline $1/K$. Otherwise, the candidate set is empty:
\begin{equation}
 \mathcal E_{ij}=\begin{cases}
 \{k:\operatorname{rank}(w_{ij}^k)\le m^*,\ w_{ij}^k>1/K\}, & \widetilde g_{ij}\ge\tau,\\
 \varnothing, & \widetilde g_{ij}<\tau.
 \end{cases}
 \label{eq:gap-gate}
\end{equation}
Multiplying the gap by $K$ expresses it relative to the uniform weight $1/K$. Thus, a fixed threshold $\tau$ corresponds to a raw-gap threshold of $\tau/K$; we use $\tau=0.5$ in the main configuration. The equations assume that all components are available. When some are unavailable, both attention normalization and gating use only the available components, whose count replaces $K$. When fewer than two components are available, reasoning tokens receive the calibrated response-level advantage. The gate selects a proper subset of the available components; with two available components, it therefore selects one component or falls back.

\paragraph{Same-sign aggregation and conflicting-sign handling.}
The candidate set determines which components contribute to the sentence credit. We use the candidate advantage directly when only one component is selected. For multiple candidates, we average their advantages if the set does not contain both a strictly positive and a strictly negative advantage. If both signs are present, averaging may cancel the advantages; we instead use the advantage of the candidate with the largest estimated association weight. The resulting credit for sentence $U_{i,j}$ is
\begin{equation}
 c_{ij}=
 \begin{cases}
 \widetilde A_i^k, & \mathcal E_{ij}=\{k\},\\
 |\mathcal E_{ij}|^{-1}\!\displaystyle\sum_{k\in\mathcal E_{ij}}\widetilde A_i^k,
 & |\mathcal E_{ij}|>1\ \text{and no conflicting-sign pair exists},\\
 \widetilde A_i^{k^*},\quad k^*=\displaystyle\arg\max_{k\in\mathcal E_{ij}}w_{ij}^k,
 & |\mathcal E_{ij}|>1\ \text{and a conflicting-sign pair exists},\\
 \widetilde A_i, & \mathcal E_{ij}=\varnothing.
 \end{cases}
 \label{eq:sentence-credit}
\end{equation}
Estimated association weights determine candidate selection, whereas the selected component-specific advantages determine the sentence credit. If the candidate set is empty, the sentence receives the calibrated response-level advantage $\widetilde A_i$.

\paragraph{Credit for three token types and the policy objective.}
After sentence credit has been computed, raw credit is assembled for three types of generated tokens: (1) field-value tokens, whose component identity is specified by the schema, receive $\widetilde A_i^k$; (2) reasoning tokens receive the credit $c_{ij}$ of their sentence; and (3) all remaining tokens retain the response-level advantage $\widetilde A_i$:
\begin{equation}
 C^{\rm raw}_{it}=\begin{cases}
 \widetilde A_i^k, & t\in S_i^k,\\
 c_{ij}, & t\in U_{i,j},\\
 \widetilde A_i, & \text{otherwise}.
 \end{cases}
 \label{eq:raw-credit}
\end{equation}
Here, $S_i^k$ contains only successfully located field-value positions for which a component-specific advantage is available. Reasoning positions that cannot be routed use $\widetilde A_i$. Failure to route a reasoning sentence does not undo field-value credit that has already been assigned. If neither field-value assignment nor reasoning routing is performed, all valid positions initially receive $\widetilde A_i$, before final whitening and clipping.

Routing changes how many tokens receive each advantage. To standardize the assembled token credit, we jointly whiten it, then clip it to $[-3,3]$. Let $m_{it}\in\{0,1\}$ denote the valid-response-token mask:
\begin{equation}
 \widehat C_{it}=m_{it}\operatorname{clip}\!\left(\operatorname{Whiten}(C^{\rm raw})_{it},-3,3\right).
 \label{eq:final-credit}
\end{equation}
Sign-aware advantage combination precedes global token-credit whitening.

We use the final credit in a clipped policy objective with token-level importance ratios. For token $t$ in response $i$, $\rho_{it}(\theta)$ is the ratio of its probabilities under the current and old policies, conditioned on the same input and preceding tokens. We replace the response-level advantage at each position with $\widehat C_{it}$:
\begin{equation}
 \mathcal L_{\mathrm{AC}}(\theta)
 =-\mathbb E_{i,t}\!\left[
 \min\!\left(
 \rho_{it}(\theta)\widehat C_{it},
 \operatorname{clip}(\rho_{it}(\theta),1-\epsilon,1+\epsilon)\widehat C_{it}
 \right)
 \right].
 \label{eq:attrcredit-loss}
\end{equation}
Equation~\eqref{eq:attrcredit-loss} shows the standard symmetric clipping form. Each implementation retains the underlying algorithm's sampling procedure, token-level importance ratios, clipping settings, and KL regularization when used. Replacing the response-level advantage with position-specific credit changes the surrogate objective.

\paragraph{Implementation.}\label{sec:implementation}
Our FlashAttention-based training stack does not expose the attention weights needed for routing. We therefore use an additional forward pass without gradients to extract anchor attention and aggregate it by reasoning sentence; no additional backward pass is required. In the profiled training interval, the attribution prepass takes 19.85 seconds per step, accounting for 13.45\% of \method{}'s total step time. Integrating anchor attention extraction into an existing training pass could reduce this overhead. Its runtime and memory trade-offs across training stacks warrant further study.

\section{Experiments}\label{sec:experiments}

We evaluate whether \method{} improves multi-component prediction, whether its benefits extend across training frameworks, and how its credit-assignment choices affect performance. We then examine sentence--component associations independently of downstream prediction metrics.

\subsection{Experimental Setup}\label{sec:exp-setup}

\paragraph{Tasks and rewards.}
ProductAttr contains 7,200 product image--text samples, split into 6,400 training and 800 evaluation examples. It requires predicting packaged weight, dimensions, and foldability, with separate rewards for the three attribute groups. On Nutrition5k~\citep{thames2021nutrition5k}, we predict fat, carbohydrates, and protein from one overhead image per dish, using 2,755 training and 507 evaluation dishes from the official RGB splits. We also evaluate dialogue state tracking on Schema-Guided Dialogue (SGD; \citealp{rastogi2020towards}). Data preparation, output fields, and evaluation conventions are detailed in Appendix~\ref{sec:appendix-data}.

\paragraph{Models and comparisons.}
Our main experiments use Qwen3.5-4B and compare the initial model without task-specific RL (No-RL), GRPO~\citep{shao2024deepseekmath}, GDPO~\citep{liu2026gdpo}, an adapted GD$^2$PO~\citep{liu2026gd2po}, and RTT~\citep{xu2026rtt}. Unless specified otherwise, \method{} is added to GRPO. In the tables, Ours denotes the \method{} module. Within each task, comparisons use the same data and evaluation protocol; host-framework integrations use matched configurations within each pair. ProductAttr training uses 600 updates with eight responses per prompt. Full training settings and the integration interface appear in Appendix~\ref{sec:appendix-training}.

\paragraph{Metrics and statistics.}
We report ratio-of-sums RMAE for continuous attributes, F1 for foldability, and Joint/Average Goal Accuracy (JGA/AGA) for SGD. Unless stated otherwise, means and sample standard deviations use three decoding seeds for a fixed training checkpoint. Independent training-seed replication is reported separately. Paired confidence intervals and their resampling units are specified in Appendix~\ref{sec:appendix-statistics}.

\subsection{Main Results}\label{sec:main-results}

\begin{table}[htbp]
\centering
\caption{Main results on ProductAttr and Nutrition5k (\%). Each method uses one fixed training checkpoint. Avg. is the arithmetic mean of fat, carbohydrate, and protein RMAE, following the macronutrient summary in \citet{thames2021nutrition5k}. The statistical protocol is specified in Section~\ref{sec:exp-setup}.}
\label{tab:main-results}
\scriptsize
\renewcommand{\tablesd}[1]{\text{\normalfont\fontsize{6}{7}\selectfont\color{black!65}\ensuremath{\pm #1}}}
\setlength{\tabcolsep}{1pt}
\begin{tabular*}{\textwidth}{@{\extracolsep{\fill}}l*{7}{c}@{}}
\toprule
& \multicolumn{3}{c}{ProductAttr-Eval} & \multicolumn{4}{c}{Nutrition5k (RMAE $\downarrow$)}\\
\cmidrule(lr){2-4}\cmidrule(lr){5-8}
Method & \shortstack{Weight\\RMAE $\downarrow$} & \shortstack{Dimensions\\RMAE $\downarrow$} & \shortstack{Fold\\F1 $\uparrow$} & Avg. $\downarrow$ & Fat & Carbs & Protein\\
\midrule
No-RL & $112.80\tablesd{16.01}$ & $81.57\tablesd{19.34}$ & $65.49\tablesd{3.77}$ & $56.33\tablesd{2.20}$ & $60.14\tablesd{3.49}$ & $55.70\tablesd{1.19}$ & $53.15\tablesd{2.04}$\\
GRPO & $30.86\tablesd{0.30}$ & $22.48\tablesd{0.76}$ & $89.55\tablesd{0.89}$ & $35.02\tablesd{0.39}$ & $40.10\tablesd{0.77}$ & $32.37\tablesd{0.41}$ & $32.59\tablesd{0.11}$\\
GDPO & $26.23\tablesd{0.31}$ & $20.54\tablesd{0.07}$ & $88.29\tablesd{0.53}$ & $38.72\tablesd{2.09}$ & $44.27\tablesd{2.18}$ & $35.12\tablesd{2.22}$ & $36.79\tablesd{2.05}$\\
GD$^2$PO & $27.38\tablesd{0.60}$ & $21.67\tablesd{0.52}$ & $78.24\tablesd{0.26}$ & $39.68\tablesd{0.48}$ & $44.91\tablesd{0.34}$ & $35.74\tablesd{0.58}$ & $38.40\tablesd{0.76}$\\
RTT & $29.87\tablesd{0.14}$ & $23.52\tablesd{0.85}$ & $90.91\tablesd{0.34}$ & $37.44\tablesd{0.53}$ & $42.27\tablesd{1.11}$ & $32.64\tablesd{0.59}$ & $37.41\tablesd{0.50}$\\
GRPO + Ours & $\bestmean{25.12}\tablesd{0.34}$ & $\bestmean{19.68}\tablesd{0.06}$ & $\bestmean{91.31}\tablesd{0.42}$ & $\bestmean{33.05}\tablesd{0.36}$ & $\bestmean{37.10}\tablesd{0.57}$ & $\bestmean{31.47}\tablesd{0.93}$ & $\bestmean{30.59}\tablesd{0.63}$\\
\bottomrule
\end{tabular*}
\end{table}

\paragraph{Multi-attribute prediction.}
Adding \method{} to GRPO improves mean performance on all three target attributes in both benchmarks (Table~\ref{tab:main-results}). On ProductAttr, weight and dimension RMAE decrease from 30.86 to 25.12 and from 22.48 to 19.68, respectively, while Fold F1 increases from 89.55 to 91.31. On Nutrition5k, the improvements extend to all three nutrients, reducing average RMAE from 35.02 to 33.05. GRPO + \method{} also achieves better mean performance than RTT on all six target attributes, without training a separate relevance discriminator. These results show that assigning component-specific credit can improve joint prediction beyond broadcasting a shared response-level advantage. Paired bootstrap intervals support the continuous-attribute gains for the fixed ProductAttr checkpoints (Appendix~\ref{sec:appendix-statistics}).

\begin{wraptable}{R}{0.53\textwidth}
\vspace{-\intextsep}\vspace{-\abovecaptionskip}
\centering\footnotesize
\setlength{\tabcolsep}{3pt}
\caption{Dialogue state tracking on SGD (\%; higher is better). Mean and sample standard deviation use three decoding seeds per fixed training checkpoint. The full comparison, including No-RL, is in Table~\ref{tab:sgd-full-results} in Appendix~\ref{sec:appendix-replication-sgd}.}
\label{tab:sgd-results}
\begin{tabular*}{\linewidth}{@{\extracolsep{\fill}}lcc@{}}
\toprule
Method & JGA $\uparrow$ & AGA $\uparrow$\\
\midrule
GRPO & $75.31\tablesd{0.06}$ & $94.48\tablesd{0.02}$\\
GRPO + Ours & $\bestmean{81.31}\tablesd{0.11}$ & $\bestmean{96.48}\tablesd{0.01}$\\
\bottomrule
\end{tabular*}
\end{wraptable}

\paragraph{Dialogue state tracking.}\label{sec:cross-domain}
The gains also extend to structured outputs in a different domain. On SGD, adding \method{} increases JGA from 75.31 to 81.31 and AGA from 94.48 to 96.48, gains of 6.00 and 2.00 percentage points (Table~\ref{tab:sgd-results}). Here, credit is assigned to dialogue-state components rather than physical attributes. This task-specific training evaluation broadens the evidence beyond image-based attribute prediction.

\paragraph{Replication across training seeds.}
The ProductAttr comparison remains favorable when summarized over three independently trained checkpoints per method. Mean weight/dimension RMAE is 26.10/20.16 with \method{}, compared with 28.35/21.21 for GRPO, and mean Fold F1 rises from 90.00 to 91.72 (Table~\ref{tab:product-training-seeds} in Appendix~\ref{sec:appendix-replication-sgd}). This replication complements the decoding-seed variation reported in the main table.

\subsection{Compatibility Across Training Frameworks and Models}\label{sec:compatibility}

\begin{table}[htbp]
\centering\footnotesize
\caption{Compatibility across training frameworks and models on ProductAttr (\%). Panel (a) integrates \method{} with DAPO and GDPO on Qwen3.5-4B; panel (b) integrates it with GRPO on larger dense and MoE models. `{}`{}+ Ours'{}'{} adds \method{} to the baseline; each pair uses matched configurations within its setting. Results follow Section~\ref{sec:exp-setup}.}
\label{tab:host-compatibility}
\setlength{\tabcolsep}{3pt}
\begin{tabular*}{\textwidth}{@{\extracolsep{\fill}}llccc@{}}
\toprule
Setting & Method & \shortstack{Weight\\RMAE $\downarrow$} & \shortstack{Dimensions\\RMAE $\downarrow$} & Fold F1 $\uparrow$\\
\midrule
\multicolumn{5}{l}{\textit{(a) Training frameworks: Qwen3.5-4B}}\\
DAPO & Original & $27.96\tablesd{1.07}$ & $21.48\tablesd{0.03}$ & $90.82\tablesd{0.13}$\\
 & + Ours & $\bestmean{26.13}\tablesd{0.42}$ & $\bestmean{20.07}\tablesd{0.03}$ & $\bestmean{91.00}\tablesd{0.21}$\\
GDPO & Original & $26.23\tablesd{0.31}$ & $20.54\tablesd{0.07}$ & $88.29\tablesd{0.53}$\\
 & + Ours & $\bestmean{25.02}\tablesd{0.41}$ & $\bestmean{19.72}\tablesd{0.18}$ & $\bestmean{91.88}\tablesd{0.43}$\\
\midrule
\multicolumn{5}{l}{\textit{(b) Model scale and architecture: GRPO}}\\
Qwen3.5-9B & GRPO & $28.17\tablesd{0.65}$ & $20.93\tablesd{0.24}$ & $91.48\tablesd{0.66}$\\
(dense) & + Ours & $\bestmean{23.49}\tablesd{0.53}$ & $\bestmean{19.22}\tablesd{0.07}$ & $\bestmean{92.79}\tablesd{0.47}$\\
Qwen3.5-35B-A3B & GRPO & $31.99\tablesd{1.27}$ & $21.35\tablesd{0.12}$ & $90.44\tablesd{0.31}$\\
(MoE) & + Ours & $\bestmean{25.06}\tablesd{0.98}$ & $\bestmean{18.94}\tablesd{0.17}$ & $\bestmean{92.25}\tablesd{0.57}$\\
\bottomrule
\end{tabular*}
\end{table}

\paragraph{Integration with DAPO and GDPO.}
To test whether the gains depend on GRPO, we integrate \method{} with DAPO~\citep{yu2025dapo} and GDPO, retaining each framework's sampling and policy-update rules. Both pairs improve all three mean ProductAttr metrics (Table~\ref{tab:host-compatibility}, panel~a). With DAPO, weight and dimension RMAE decrease by 1.83 and 1.41 percentage points. With GDPO, they decrease by 1.21 and 0.82 points, while Fold F1 increases by 3.59 points. These paired comparisons support \method{} as a reusable credit-assignment module within the evaluated frameworks. In particular, the GDPO integration shows that assigning credit to relevant positions can complement component-wise reward normalization.

\paragraph{Model and output extensions.}\label{sec:stability}
Gains also extend to larger Qwen3.5-9B and 35B-A3B MoE models (Table~\ref{tab:host-compatibility}b) and the five-attribute extension (Table~\ref{tab:product-fivetask}, Appendix~\ref{sec:appendix-product-results}). Beyond Qwen, on Gemma3-4B-IT (alternating local and global attention), parameter-free layer averaging reduces dimension RMAE from 133.86 to 50.88 and raises Fold F1 from 55.91 to 59.15 (Table~\ref{tab:scale-replication}, Appendix~\ref{sec:cross-model-results}; frozen-response layer-group diagnostics on Qwen3.5-4B and Gemma4 E4B in Appendix~G.2).

\subsection{Credit-Assignment Analysis}\label{sec:routing-validation}

\begin{wraptable}{r}{0.55\textwidth}
\vspace{-\intextsep}\vspace{-\abovecaptionskip}
\centering
\caption{Credit-assignment regions on ProductAttr-Eval (\%). Results use three evaluation seeds and the statistical protocol in Section~\ref{sec:exp-setup}.}
\label{tab:attribution-controls}
{\renewcommand{\arraystretch}{1.08}
\footnotesize\setlength{\tabcolsep}{2pt}\begin{tabular*}{\linewidth}{@{\extracolsep{\fill}}lccc@{}}
\toprule
Variant & \shortstack{Weight\\RMAE $\downarrow$} & \shortstack{Dimensions\\RMAE $\downarrow$} & \shortstack{Fold\\F1 $\uparrow$}\\
\midrule
GRPO & $30.86\tablesd{0.30}$ & $22.48\tablesd{0.76}$ & $89.55\tablesd{0.89}$\\
Value-only & $30.80\tablesd{1.12}$ & $21.88\tablesd{0.09}$ & $89.39\tablesd{0.57}$\\
Reasoning-only & $25.27\tablesd{0.72}$ & $19.90\tablesd{0.29}$ & $\bestmean{92.54}\tablesd{0.53}$\\
\midrule
\method{} & $\bestmean{25.12}\tablesd{0.34}$ & $\bestmean{19.68}\tablesd{0.06}$ & $91.31\tablesd{0.42}$\\
\bottomrule
\end{tabular*}}
\end{wraptable}

\paragraph{Where should component-specific credit be assigned?}
We compare assigning component-specific credit only to output values, only to reasoning sentences, or to both (Table~\ref{tab:attribution-controls}). Reasoning-only yields lower weight and dimension errors than Value-only, indicating that the allocation of credit within reasoning affects predictive performance. Assigning credit to both regions gives the lowest continuous errors in this comparison, while Reasoning-only gives the highest Fold F1. The lower continuous errors relative to Value-only support assigning component-specific feedback to reasoning sentences.

\paragraph{Routing granularity.}
On ProductAttr, sentence-level routing reduces weight RMAE from 27.01 to 25.12 and dimension RMAE from 20.40 to 19.68. Token-level routing yields slightly higher Fold F1 (91.46 vs.\ 91.31). Table~\ref{tab:routing-granularity} reports the full comparison.

\subsection{Offline Association Analysis}\label{sec:offline-association}

\begin{wraptable}{r}{0.49\textwidth}
\vspace{-\intextsep}\vspace{-\abovecaptionskip}
\centering\footnotesize
\caption{\small Association diagnostics on ProductAttr against Terra reference labels (\%). Multi-label association micro-F1 on Base and GPT responses. Both response sources are replayed through Qwen3.5-4B for attention and gradient extraction. Random controls average ten seeds.}
\label{tab:offline-association-summary}
\setlength{\tabcolsep}{3pt}
\begin{tabular*}{\linewidth}{@{\extracolsep{\fill}}lcc@{}}
\toprule
\multicolumn{3}{@{}l@{}}{\shortstack[l]{\textit{Multi-label association}\\\textit{Micro-F1 $\uparrow$}}}\\
Method & Base & GPT\\
\midrule
Matched random & 34.14 & 33.44\\
Keyword heuristic & 61.09 & 64.43\\
Gradient gate & 38.05 & 48.74\\
\midrule
\method{} & \bestmean{76.50} & \bestmean{73.62}\\
\bottomrule
\end{tabular*}
\end{wraptable}

\paragraph{Evaluation protocol.}
We evaluate associations on responses generated by the Qwen3.5-4B base model and GPT-5, independently of our GRPO/\method{} training runs. Each source contains eight responses for each of 128 training products. Both sources are replayed through Qwen3.5-4B to extract attention and gradient signals; GPT-generated responses do not provide GPT's internal attention. We evaluate agreement with sentence--component reference labels generated by GPT-5.6 Terra.

\WFclear

\paragraph{Association quality.}
\method{} achieves higher micro-F1 than keyword, gradient, and matched-random controls on both response sources (Table~\ref{tab:offline-association-summary}). Relative to keyword matching, F1 improves by 15.41 points on Base responses and 9.19 points on GPT responses. On Base responses, this reflects higher recall at lower precision; on GPT responses, both increase. Thus, output-anchor attention captures associations beyond explicit keyword overlap on these two sources. Appendix~\ref{sec:model-reference-association} reports precision, recall, coverage, and an ungated Top-1 variant, which routes every sentence. The ungated variant has higher F1, highlighting the coverage trade-off introduced by candidate gating.

\begin{wraptable}{r}{0.49\textwidth}
\vspace{-\intextsep}
\centering\small
\vspace{-\abovecaptionskip}
\caption{Top-1 primary-component precision on GPT responses (\%). Precision measures the fraction of predicted assignments matching the reference; predictions on empty-reference sentences count as errors.}
\label{tab:primary-component-precision}
\setlength{\tabcolsep}{3pt}
\begin{tabular*}{\linewidth}{@{\extracolsep{\fill}}lc@{}}
\toprule
\multicolumn{2}{@{}l@{}}{\shortstack[l]{\textit{Primary-component prediction}\\\textit{GPT responses}}}\\
Method & Precision $\uparrow$\\
\midrule
Random & 32.63\\
Matched random & 32.93\\
Keyword & 90.21\\
Gradient & 41.92\\
\midrule
\method{} (gated) & \bestmean{92.29}\\
\method{} (ungated) & 84.28\\
\bottomrule
\end{tabular*}
\end{wraptable}

\paragraph{Primary-component identification.}
We additionally evaluate Top-1 primary-component prediction on GPT-generated responses (Table~\ref{tab:primary-component-precision}). The gated variant achieves 92.29\% precision, compared with 90.21\% for keyword matching, 41.92\% for gradient attribution, and 32.63\% for random assignment. Without gating, \method{} achieves 84.28\% precision. Gating retains the same Top-1 choice on accepted sentences, selecting more reliable assignments through abstention. Reference-label construction and the scoring protocol are detailed in Appendix~\ref{sec:primary-component-protocol}.

\WFclear

\section{Conclusion and Limitations}\label{sec:preregistered-identity}

\paragraph{Conclusion.}
We identified cross-component credit misassignment in structured generation, where a shared response-level advantage can oppose a component-specific learning signal. To address this mismatch, we introduced \method{}, a plug-in module that uses the policy model's attention from output-value anchors to assign component-specific credit to reasoning sentences, without additional relevance annotations or a separate relevance discriminator. To evaluate this approach, we constructed ProductAttr, a benchmark of 7,200 product image--text samples with component-wise verifiers. Experiments on this benchmark, Nutrition5k, and dialogue state tracking show improvements over GRPO. Further gains with DAPO and GDPO and on larger dense and MoE models support the module's applicability across training algorithms and model configurations.

\paragraph{Limitations.}
\method{} requires separately verifiable components and locatable output values. Our FlashAttention integration uses a no-gradient forward pass for anchor attention (profiled in Appendix~G.3). While we evaluate Qwen3.5 (4B--35B-A3B) and Gemma3-4B-IT, scaling online multimodal RLVR to 100B+ models remains future work.

\section*{Statement on the Use of AI}
We used generative AI tools for literature retrieval, experimental-method review, code implementation and debugging, statistical-analysis scripts, interpretation of results, and drafting and editing manuscript text and structure. We used GPT-5 to generate responses and GPT-5.6 Terra to annotate sentence--component associations for the offline evaluation. Qwen and GPT models assisted with ProductAttr screening and semantic-label checks, with human review of disputed cases. Product images and text come from real product records. Reported numerical results are computed from saved model outputs using fixed evaluation scripts, and important code changes are checked through tests or artifact-level verification. The authors are responsible for the final text, code, results, citations, and conclusions.

\section*{Reproducibility Statement}
The problem formulation and core algorithm are described in Section~\ref{sec:problem-formulation} and the method section. Data splits, output schemas, training configurations, statistical tests, and completed ablations are documented in the corresponding appendix sections. The AnchorCredit implementation, ROLL integration, and example training and evaluation scripts are available at \url{https://github.com/AlibabaResearch/anchorcredit}.

\section*{Ethics Statement}
Nutrition5k uses public data. Permission to release ProductAttr images, text, labels, fixed splits, and accompanying code has been confirmed. ProductAttr has not yet been publicly released. The planned release will follow the applicable licenses and usage terms and exclude credentials, internal infrastructure details, and sensitive fields.

\bibliographystyle{plainnat}
\bibliography{references}

@inproceedings{rastogi2020towards,
  title     = {Towards Scalable Multi-Domain Conversational Agents: The Schema-Guided Dialogue Dataset},
  author    = {Rastogi, Abhinav and Zang, Xiaoxue and Sunkara, Srinivas and Gupta, Raghav and Khaitan, Pranav},
  booktitle = {Proceedings of the AAAI Conference on Artificial Intelligence},
  volume    = {34},
  number    = {05},
  pages     = {8689--8696},
  year      = {2020},
  doi       = {10.1609/aaai.v34i05.6394},
  url       = {https://ojs.aaai.org/index.php/AAAI/article/view/6394}
}

@inproceedings{lightman2024verify,
  title = {Let's Verify Step by Step},
  author = {Lightman, Hunter and Kosaraju, Vineet and Burda, Yura and Edwards, Harri and Baker, Bowen and Lee, Teddy and Leike, Jan and Schulman, John and Sutskever, Ilya and Cobbe, Karl},
  booktitle = {International Conference on Learning Representations},
  year = {2024},
  url = {https://proceedings.iclr.cc/paper_files/paper/2024/file/aca97732e30bcf1303bc22ac3924fd16-Paper-Conference.pdf}
}

@inproceedings{chan2024dense,
  title = {Dense Reward for Free in Reinforcement Learning from Human Feedback},
  author = {Chan, Alex James and Sun, Hao and Holt, Samuel and van der Schaar, Mihaela},
  booktitle = {Proceedings of the 41st International Conference on Machine Learning},
  series = {Proceedings of Machine Learning Research},
  volume = {235},
  pages = {6136--6154},
  year = {2024},
  url = {https://proceedings.mlr.press/v235/chan24a.html}
}

@article{xie2026stepwise,
  title = {Step-wise Rubric Rewards for {LLM} Reasoning},
  author = {Xie, Weichu and Zhao, Haozhe and Liu, Wenpu and Zhu, Yongfu and Chen, Liang and Ye, Minghao and Chen, Zirong and Xu, Yuqi and Dong, Shuai and Wang, Ziyue and Xu, Xinbo and Shi, Kean and Wu, Ruoyu and Zhang, Xiaoying and Shao, Wenqi and Chang, Baobao and Duan, Nan and Wang, Jiaqi},
  journal = {arXiv preprint arXiv:2605.17291},
  year = {2026},
  url = {https://arxiv.org/abs/2605.17291}
}

@article{pavlenko2026blockwise,
  title   = {Blockwise Advantage Estimation for Multi-Objective {RL} with Verifiable Rewards},
  author  = {Pavlenko, Kirill and Golubev, Alexander and Karasik, Simon and Yangel, Boris},
  journal = {arXiv preprint arXiv:2602.10231},
  year    = {2026},
  url     = {https://arxiv.org/abs/2602.10231}
}

@article{yang2021mave,
  title   = {{MAVE}: A Product Dataset for Multi-source Attribute Value Extraction},
  author  = {Yang, Li and Wang, Qifan and Yu, Zac and Kulkarni, Anand and Sanghai, Sumit and Shu, Bin and Elsas, Jon and Kanagal, Bhargav},
  journal = {arXiv preprint arXiv:2112.08663},
  year    = {2021}
}

@article{zhai2024physical,
  title   = {Physical Property Understanding from Language-Embedded Feature Fields},
  author  = {Zhai, Albert J. and Shen, Yuan and Chen, Emily Y. and Wang, Gloria X. and Wang, Xinlei and Wang, Sheng and Guan, Kaiyu and Wang, Shenlong},
  journal = {arXiv preprint arXiv:2404.04242},
  year    = {2024}
}

@inproceedings{thames2021nutrition5k,
  title     = {{Nutrition5k}: Towards Automatic Nutritional Understanding of Generic Food},
  author    = {Thames, Quin and Karpur, Arjun and Norris, Wade and Xia, Fangting and Panait, Liviu and Weyand, Tobias and Sim, Jack},
  booktitle = {Proceedings of the IEEE/CVF Conference on Computer Vision and Pattern Recognition},
  pages     = {8903--8911},
  year      = {2021}
}

@article{shao2024deepseekmath,
  title   = {{DeepSeekMath}: Pushing the Limits of Mathematical Reasoning in Open Language Models},
  author  = {Shao, Zhihong and Wang, Peiyi and Zhu, Qihao and Xu, Runxin and Song, Junxiao and Zhang, Mingchuan and Li, Y. K. and Wu, Y. and Guo, Daya},
  journal = {arXiv preprint arXiv:2402.03300},
  year    = {2024}
}

@article{deepseekai2025r1,
  title   = {{DeepSeek-R1}: Incentivizing Reasoning Capability in {LLMs} via Reinforcement Learning},
  author  = {{DeepSeek-AI}},
  journal = {arXiv preprint arXiv:2501.12948},
  year    = {2025}
}

@article{yu2025dapo,
  title   = {{DAPO}: An Open-Source {LLM} Reinforcement Learning System at Scale},
  author  = {Yu, Qiying and Zhang, Zheng and Zhu, Ruofei and others},
  journal = {arXiv preprint arXiv:2503.14476},
  year    = {2025}
}

@article{liu2026gdpo,
  title   = {{GDPO}: Group Reward--Decoupled Normalization Policy Optimization for Multi-Reward {RL} Optimization},
  author  = {Liu, Shih-Yang and Dong, Xin and Lu, Ximing and Diao, Shizhe and Belcak, Peter and Liu, Mingjie and Chen, Min-Hung and Yin, Hongxu and Wang, Yu-Chiang Frank and Cheng, Kwang-Ting and Choi, Yejin and Kautz, Jan and Molchanov, Pavlo},
  journal = {arXiv preprint arXiv:2601.05242},
  year    = {2026}
}

@article{liu2026gd2po,
  title   = {{GD$^2$PO}: Mitigating Multi-Reward Conflicts via Group-Dynamic Reward-Decoupled Policy Optimization},
  author  = {Liu, Haotian and Liu, Yihao and Ni, Jingwei and Huang, Siyuan and Liu, Xinpeng and Cheng, Pengyu and Song, Jiajun and Ding, Ruijin and Li, Junfeng and Yu, Zhechao and Zhou, Mengyu and Xu, Hongteng and Jiang, Xiaoxi and Jiang, Guanjun},
  journal = {arXiv preprint arXiv:2606.16771},
  year    = {2026}
}

@article{li2025attention,
  title   = {Attention Illuminates {LLM} Reasoning: The Preplan-and-Anchor Rhythm Enables Fine-Grained Policy Optimization},
  author  = {Li, Yang and Dong, Zhichen and Sun, Yuhan and Wang, Weixun and Xiong, Shaopan and Luo, Yijia and Liu, Jiashun and Lu, Han and Wang, Jiamang and Su, Wenbo and Zheng, Bo and Yan, Junchi},
  journal = {arXiv preprint arXiv:2510.13554},
  year    = {2025}
}

@article{dong2026flowtracer,
  title   = {How Does Reasoning Flow? Tracing Attention-Induced Information Flow for Targeted {RL} in {LLMs}},
  author  = {Dong, Zhichen and Li, Yang and Sun, Yuhan and Wang, Weixun and Luo, Yijia and Peng, Zinian and Ye, Taiheng and Yang, Chao and Su, Wenbo and Cheng, Yu and Zheng, Bo and Yan, Junchi},
  journal = {arXiv preprint arXiv:2606.10646},
  year    = {2026}
}

@inproceedings{xu2026rtt,
  title     = {Rubrics to Tokens: Bridging Response-Level Rubrics and Token-Level Rewards in Instruction Following Tasks},
  author    = {Xu, Tianze and Zheng, Yanzhao and Lu, Pengrui and Ye, Lyumanshan and Wu, Yong and Zhang, Zhentao and Yu, Yuanqiang and Ma, Chao and Zhu, Jihuai and Liu, Pengfei and Dong, Baohua and Zhu, Hangcheng and Huang, Ruohui and Yu, Gang},
  booktitle = {Conference on Language Modeling},
  year      = {2026},
  url       = {https://arxiv.org/abs/2604.02795}
}

@inproceedings{wang2024mathshepherd,
  title     = {{Math-Shepherd}: Verify and Reinforce {LLM}s Step-by-step without Human Annotations},
  author    = {Wang, Peiyi and Li, Lei and Shao, Zhihong and Xu, Runxin and Dai, Damai and Li, Yifei and Chen, Deli and Wu, Yu and Sui, Zhifang},
  booktitle = {Proceedings of the 62nd Annual Meeting of the Association for Computational Linguistics},
  pages     = {9426--9439},
  year      = {2024},
  doi       = {10.18653/v1/2024.acl-long.510}
}

@inproceedings{setlur2025rewarding,
  title     = {Rewarding Progress: Scaling Automated Process Verifiers for {LLM} Reasoning},
  author    = {Setlur, Amrith and Nagpal, Chirag and Fisch, Adam and Geng, Xinyang and Eisenstein, Jacob and Agarwal, Rishabh and Agarwal, Alekh and Berant, Jonathan and Kumar, Aviral},
  booktitle = {International Conference on Learning Representations},
  year      = {2025},
  url       = {https://openreview.net/forum?id=A6Y7AqlzLW}
}

@inproceedings{kazemnejad2025vineppo,
  title     = {{VinePPO}: Refining Credit Assignment in {RL} Training of {LLM}s},
  author    = {Kazemnejad, Amirhossein and Aghajohari, Milad and Portelance, Eva and Sordoni, Alessandro and Reddy, Siva and Courville, Aaron and Le Roux, Nicolas},
  booktitle = {Proceedings of the 42nd International Conference on Machine Learning},
  volume    = {267},
  series    = {Proceedings of Machine Learning Research},
  pages     = {29557--29590},
  year      = {2025}
}

@inproceedings{guo2025spo,
  title     = {Segment Policy Optimization: Effective Segment-Level Credit Assignment in {RL} for Large Language Models},
  author    = {Guo, Yiran and Xu, Lijie and Liu, Jie and Ye, Dan and Qiu, Shuang},
  booktitle = {Advances in Neural Information Processing Systems},
  year      = {2025}
}

@inproceedings{jain2019attention,
  title     = {Attention is not Explanation},
  author    = {Jain, Sarthak and Wallace, Byron C.},
  booktitle = {Proceedings of NAACL-HLT},
  pages     = {3543--3556},
  year      = {2019},
  doi       = {10.18653/v1/N19-1357}
}

@inproceedings{wiegreffe2019attention,
  title     = {Attention is not not Explanation},
  author    = {Wiegreffe, Sarah and Pinter, Yuval},
  booktitle = {Proceedings of EMNLP-IJCNLP},
  pages     = {11--20},
  year      = {2019},
  doi       = {10.18653/v1/D19-1002}
}

\clearpage
\appendix

\section{Data, Prompts, and Verifiers}\label{sec:appendix-data}

\subsection{Data Splits and Deduplication}

\paragraph{Product sources and labels.}
ProductAttr contains 7,200 distinct AliExpress products, each represented by one stock-keeping unit (SKU), spanning 26 top-level categories, 100 second-level categories, and 1,075 leaf categories. Product images and text are matched to warehouse records by product and SKU identifiers. Weight and dimensions refer to the packaged product, measured in grams and millimeters, rather than net weight or unpackaged dimensions. For repeated measurements, the maximum recorded values are retained. All physical labels are complete, parseable, and positive.\par Foldability indicates whether the SKU can be safely folded or rolled under fulfillment rules that account for packaging and damage risk. The dataset contains 5,040 nonfoldable and 2,160 foldable products. The supplementary seasonality labels indicate pronounced seasonal relevance in temperate Northern Hemisphere markets and may include multiple seasons. Festival labels identify one of seven specific holidays from the product's design or intended use; products without a clear holiday association receive the nonfestive label. Semantic labels are checked against product images, text, and business rules using Qwen and GPT models, with human review of disputed cases.\par Construction starts from 100 high-sales categories. Model-assisted screening removes inconsistent product information, image--text mismatches, category errors, and implausible warehouse measurements. Additional candidates are retrieved to improve coverage of rare festival labels. After label review and product-level deduplication, deterministic sampling maintains category coverage and increases representation of rare labels. The resulting label proportions reflect dataset construction rather than the product population.

\paragraph{Input attributes and quality checks.}
Standard product attributes are obtained from the platform's product-information system. Only item-level standard English attribute names and values are retained, separately from SKU-selection attributes. When constructing the input attribute list, we remove weight, dimensions, foldability, seasonality, and festival fields, including aliases and direct proxies, as well as identifiers and backend metadata. This filtering applies only to the attribute list; natural evidence in titles, descriptions, and images is retained. After cleaning, 6,714 samples have nonempty standard attributes and 486 have empty attribute lists. Automated checks cover field structure, valid enumerations, physical-value validity, and explicit label fields in the attribute list; they do not replace manual quality checks for each product.

The dataset retains the existing seed-42 split of 6,400 training samples and 800 ProductAttr-Eval samples. Before splitting, image URLs, titles, and descriptions undergo Unicode NFKC, case, and whitespace normalization. Samples sharing any nonempty key are connected into the same group. There are 6,310 groups, with at most 70 samples in a group, and none of these exact-duplicate keys overlap across splits. After grouping, distributions are balanced over category, foldability, currency, text-missingness status, and weight, dimension, and volume bins. Deduplication uses normalized exact-match keys.

Images are snapshots of the original product images. All methods share the same preprocessing and data splits. Table~\ref{tab:appendix-data-split} summarizes the data splits and release scope.

\begin{table}[H]
\centering
\small
\caption{Data splits, deduplication, and release scope.}
\label{tab:appendix-data-split}
\begin{tabularx}{\textwidth}{lrrrX}
\toprule
Dataset & Train & Eval & Total examples & Grouping and release scope \\
\midrule
ProductAttr & 6,400 & 800 & 7,200 & Exact-duplicate groups; planned release of images, text, and labels \\
Nutrition5k & 2,755 & 507 & 3,262 & Official dish IDs; public data \\
\bottomrule
\end{tabularx}
\end{table}

\paragraph{Product benchmark release scope and status.}
Release of the complete product images, product text, attribute labels, fixed splits, and accompanying code has been confirmed as permissible. We plan to release the ProductAttr benchmark separately with documentation and license information.

The Nutrition5k-Overhead-507 adaptation uses 2,755/507 dishes from the official RGB train/test dish-ID lists. The two dish-ID sets are disjoint and contain 3,262 dishes in total. Training constructs six deterministic image variants for each training dish, yielding 16,530 training examples. These variants are not counted as independent dishes and never cross the train/test boundary. The adaptation does not introduce a separate validation split. All methods use a fixed 517-step budget and evaluate their final checkpoints on the fixed evaluation set of 507 dishes. At evaluation, each dish uses one overhead RGB image aligned with official metadata by dish ID. This setting is not a direct reproduction of the official multiview 2D main protocol. Final materials will release the complete dish-ID lists, their correspondence to the official splits, the single-image selection rule, and fixed resizing and normalization procedures. The formal task predicts fat, carbohydrates, and protein, reporting RMAE separately for the three attributes. Paired methods use the same dish lists and preprocessing. Dishes with zero-valued targets remain in the fixed evaluation set; their absolute errors contribute to the numerator of the dataset-level ratio-of-sums RMAE defined in Appendix~\ref{sec:appendix-eval}, which does not divide by an individual dish's target.

\subsection{Output Fields and Verifiers}

The product task is trained with three reward/credit groups: WT (weight), DIM (joint dimensions), and F (foldability). Five output fields do not constitute five independent tasks. Length, width, and height share one attribute reward and one routing group (Table~\ref{tab:appendix-verifier}).

\begin{table}[H]
\centering
\small
\caption{Output fields and the three attribute groups in the main product setting.}
\label{tab:appendix-verifier}
\begin{tabular}{llll}
\toprule
Attribute group & JSON field & Unit/type & Verifier \\
\midrule
Product WT & \texttt{item\_weight\_cn} & g & Weight scoring \\
Product DIM & \begin{tabular}[t]{@{}l@{}}\texttt{item\_length\_cn}, \texttt{item\_width\_cn},\\\texttt{item\_height\_cn}\end{tabular} & mm & Joint dimension scoring \\
Product F & \texttt{final\_fold\_flag} & 0/1 & Label matching \\
\bottomrule
\end{tabular}
\end{table}

Nutrition5k predicts fat, carbohydrates, and protein, with three separately reported RMAE values.

\paragraph{ProductAttr training rewards.}
Let $p$ and $g$ denote a prediction and its target. Weight is measured in grams and dimensions in millimeters. For $0<g<500$, the weight reward combines a 50-gram bucket match, distance to the target bucket, and a continuous log-ratio score. Define $b(x)=\lceil x/50\rceil$, the target interval $I_g=[\max\{0,50(b(g)-1)\},50b(g)]$, and $d(p,I_g)$ as the distance from $p$ to this interval. With $H=\mathbf{1}[p>0\text{ and }b(p)=b(g)]$ and $C=\max\{0,1-|\ln(p/g)|/\ln 4\}$ for $p>0$ ($C=0$ otherwise), the weight reward is
\[
 r_{\mathrm{WT}}=
 \begin{cases}
  0.60H+0.25\exp\!\left(-\dfrac{d(p,I_g)}{50\times0.5}\right)+0.15C, & 0<g<500,\\[3pt]
  \max\!\left\{0,1-\dfrac{|p-g|}{|g|+10^{-6}}\right\}, & g\ge500.
 \end{cases}
\]
The three dimension fields contribute to one reward:
\[
 E_{\mathrm{DIM}}=
 \frac{\sum_{a\in\{L,W,H\}}|p_a-g_a|}{\sum_{a\in\{L,W,H\}}|g_a|+10^{-6}},
 \qquad r_{\mathrm{DIM}}=\max\{0,1-E_{\mathrm{DIM}}\}.
\]
This training reward compares the original named axes. The implementation separately constructs the dimension signal used as input to component-specific advantage estimation: its bucket correction sorts the three axes and uses 50-mm buckets, a correction ratio of 0.2, and a correction cap of 0.05. These settings affect \texttt{adv\_signal\_dimensions}; the scalar reward uses the WMAPE-based $r_{\mathrm{DIM}}$ above.

Foldability receives reward 1 when prediction and target match after conversion to strings, whitespace stripping, and lowercasing, and 0 otherwise. A missing foldability prediction receives 0. With all labels present, the response reward is $R=0.4r_{\mathrm{WT}}+0.4r_{\mathrm{DIM}}+0.2r_{\mathrm{F}}$. A JSON parse failure gives reward 0 for each labeled component. Missing numerical predictions enter the corresponding reward formula as zero. Components with missing labels are excluded, and the aggregation weights are renormalized over the remaining components. The main configuration disables the additional \method{} premise penalty.

\paragraph{Nutrition5k training rewards.}
The three active fields are \texttt{fat\_g}, \texttt{carb\_g}, and \texttt{protein\_g}. For positive predictions and targets, the component reward and response reward are
\[
 r_k=\max\{0,1-|\log_2(p_k/g_k)|\},
 \qquad R=\frac{r_{\mathrm{fat}}+r_{\mathrm{carb}}+r_{\mathrm{protein}}}{3}.
\]
Thus, an exact prediction receives 1, whereas a prediction at half or twice the target receives 0. Missing, nonnumerical, nonfinite, or negative predictions receive component reward 0 and remain in the three-component denominator. A zero prediction receives 0 for a positive target; when the target is zero, only a zero prediction receives 1. JSON parse failures or a key set differing from the three active fields receive format reward 0. Severe repetition sets the response reward and all three component rewards to $-0.4$. It is triggered by a substantive sentence occurring at least five times, or by 20-character segments that each occur at least three times and together cover at least 25\% of the reasoning text. The configuration applies no additional short-reasoning, thinking-tag, or soft length penalty.

\subsection{Evaluation Metrics}\label{sec:appendix-eval}

Ratio-of-sums RMAE for each reported numerical output in either task is
\[
 R_k=\frac{\sum_{i=1}^{N}|\hat y_{ik}-y_{ik}|}{\sum_{i=1}^{N}|y_{ik}|}.
\]
For Nutrition5k, $k$ indexes fat, carbohydrates, or protein. A dish with $y_{ik}=0$ remains in the evaluation set: its absolute error contributes to the numerator, while the dataset-level denominator avoids per-dish division by zero. The numerator and denominator of joint-dimension error $R_{\mathrm{DIM}}$ each sum over both samples and the three axes. Fold F1 is reported separately.

\subsection{Prompts and Structured Outputs}

ProductAttr uses a frozen English task prompt, while product text retains its original language. Each message supplies one product image together with the title, description, top-level/second-level/leaf categories, standard attributes, SKU attributes, and SKU price with currency. Supervision labels are not provided to the model. The complete template will be released with the dataset version. Its output constraint reads:
\begin{quote}
\begin{minipage}{\linewidth}
\small
First provide your analysis as plain text. Then finish the response with exactly one JSON object containing exactly these five keys:
\par\medskip
{\ttfamily
\{"item\_weight\_cn": 0.0,\\
\hspace*{1em}"item\_length\_cn": 0.0,\\
\hspace*{1em}"item\_width\_cn": 0.0,\\
\hspace*{1em}"item\_height\_cn": 0.0,\\
\hspace*{1em}"final\_fold\_flag": 0\}}
\end{minipage}
\end{quote}
The example zero values illustrate field types and are not target labels. The template requires positive finite continuous values, Fold in 0/1, no units in numerical values, no Markdown code fence around the JSON, and no content after the final JSON. It also requires the model to distinguish whether numerical values in the description refer to the current SKU, a component, another specification, or shipping packaging, and to use the image to judge applicability. The prompt does not require brief reasoning or enforce \texttt{<think>} or \texttt{<answer>} tags.

\section{Training Configurations and Reproducibility Details}\label{sec:appendix-training}

Table~\ref{tab:appendix-training-config} lists the main ProductAttr training settings. Each method changes only its prescribed advantage or credit construction; host-framework pairs retain their native rules.

\begin{table}[H]
\centering
\small
\setlength{\tabcolsep}{4pt}
\caption{Training and generation settings. The main comparison uses training seed~42. Core product methods are also trained with seeds~43 and~44; see the multiple-training-seed results table.}
\label{tab:appendix-training-config}
\begin{tabular}{ll}
\toprule
Setting & Main ProductAttr setting \\
\midrule
Model/initialization & Qwen3.5-4B / Base \\
Learning rate and KL & $10^{-6}$; KL loss $2\times10^{-3}$ \\
clip / loss aggregation & $0.20$ / seq-mean-token-mean \\
rollout / effective batch & 32 prompts $\times$ 8 = 256 \\
Input/output limits & 5,120 / 8,192 tokens \\
Training $T$/top-$p$/top-$k$ & $0.9/0.99/100$ \\
Evaluation sampling & $0.9/0.99/100$ \\
evaluation seeds & 42, 43, 44 \\
Update budget/endpoint & 600 / checkpoint-599 \\
Checkpoint interval & 100 updates and final checkpoint \\
Hardware & 8 A100 GPUs per run \\
\bottomrule
\end{tabular}
\end{table}

ProductAttr uses AdamW with weight decay 0.01, a 25-step warmup, and a cosine learning-rate schedule. Training uses one PPO epoch, a frozen visual module, and bfloat16 (bf16). The main configuration uses gap threshold $\tau=0.50$; thresholds of 0.75 and 1.00 are evaluated in Table~\ref{tab:product-gap}.

\paragraph{\method{} pseudocode.}
Token credit is constructed in the following order:
\begin{enumerate}
  \item Standardize rewards separately for each output within each prompt's rollout group to obtain $A_i^k$, and obtain $A_i$ from the total reward. Whiten these quantities to obtain $\widetilde A_i^k$ and $\widetilde A_i$.
  \item Locate the output anchors $S_i^k$ and assign $\widetilde A_i^k$ to their tokens.
  \item Split reasoning into sentences $U_{i,j}$, aggregate attention from anchors to each sentence, and normalize over available outputs to obtain $w_{ij}^k$.
  \item Apply the normalized largest-gap gate to the sorted $w_{ij}^k$ to obtain candidate set $\mathcal E_{ij}$. Reasoning sentences that fail the gate use $\widetilde A_i$.
  \item For a single candidate, use its $\widetilde A_i^k$. For multiple candidates without conflicting signs, average their advantages. For candidates with conflicting signs, select the output-specific advantage with the largest association weight.
  \item Use $\widetilde A_i^k$ for the corresponding field-value tokens, $c_{ij}$ for reasoning-sentence tokens, and $\widetilde A_i$ for other tokens. Whiten and clip valid tokens again to obtain $\widehat C_{it}$, and insert this credit into the underlying clipped policy objective.
\end{enumerate}

\paragraph{Two-stage whitening.}
The policy surrogate uses group-relative advantage standardization, pre-routing minibatch whitening, fallback, policy-ratio clipping, and post-assembly whole-batch whitening. The first whitening operation separately calibrates the batch scales of each output-advantage column and the response-level advantage. The second jointly centers and scales all valid tokens after field, sentence, and fallback credit have been assembled. Consequently, the second operation also changes final coefficients at unrouted positions. Section~\ref{sec:routing-validation} describes the Value-only versus GRPO field-credit comparison.

\paragraph{Integration across algorithms.}
The common \method{} interface receives output-specific reward vectors from the same rollout group and the policy statistics required by the underlying policy-optimization algorithm. It returns assembled raw credit $C^{\rm raw}_{it}$ and final credit $\widehat C_{it}$. In the main GRPO experiment, $\widehat C_{it}$ replaces the response-level advantage while sampling, clipping, and regularization remain unchanged. DAPO can use the same token-credit interface, but its results are interpreted only within separately matched runs. The GDPO integration first standardizes individual rewards according to GDPO's rules and computes the whole-batch mean and standard deviation of the aggregated advantage. Contributions from each output are centered and scaled using that same mean and standard deviation so that their average agrees with GDPO's response-level advantage. The contributions are then assigned to relevant positions and final token whitening is applied. This integration does not recompute a separate batch mean and standard deviation for each output advantage. Cross-optimizer results are interpreted only within their matched controls; absolute scores do not replace the GRPO-based method comparison in the main text.

\paragraph{Reasoning-sentence segmentation.}
Sentence boundaries follow sentence-final punctuation and blank lines, excluding decimal points.

\paragraph{Attention aggregation.}
The implementation equally averages all full-attention layers that can return explicit attention matrices. It further averages over all attention heads, value tokens for the same output, and tokens within the same reasoning sentence. Linear-attention layers in hybrid architectures that do not produce such matrices are excluded. Averaging all available layers introduces no additional layer-selection hyperparameter.

\subsection{RTT Adaptation}\label{sec:appendix-rtt}

We adapt RTT~\citep{xu2026rtt} using a separately trained Qwen3.5-4B discriminator with a scalar token-classification head. It takes an attribute criterion and a generated response as input and outputs a relevance probability for each response token. For ProductAttr, supervision covers weight, joint dimensions, and foldability. The labeling pipeline combines \texttt{qwen3-max} relevance spans with deterministic JSON-field spans and negative examples in which the criterion is omitted. It uses 608 responses from 76 policy-training entities, split into 60 training and 16 validation entities, yielding 2,880 training and 768 validation criterion--response pairs. For Nutrition5k, LLM-labeled spans from training-set responses are converted to character-level labels and split by dish. Its discriminator covers five nutrient attributes, while the Macro3 policy queries only fat, carbohydrates, and protein. Discriminator training and validation use neither policy-validation nor test examples.

The discriminator uses binary cross-entropy on labeled response tokens, masking prompts, padding, and invalid labels. The training entry point specifies Qwen3.5-4B initialization, two epochs, learning rate $9\times10^{-6}$, maximum length 2,048, global batch size 64, and microbatch size 1 per GPU on four GPUs, with seed 42, bf16, ZeRO-2, SDPA, and gradient checkpointing. It is pinned to official RTT revision \texttt{b1ab2fba9bec}. During policy training, the discriminator supplies continuous relevance probabilities without binary thresholding. The policy and discriminator check vocabulary token-ID compatibility and reuse the generated token IDs for alignment.

For response $i$, component $k$, and token $t$, let $z_{ikt}$ be the discriminator logit. We compute $q_{ikt}=\operatorname{sigmoid}(z_{ikt})$ and $u_{ikt}=q_{ikt}r_i^k$. The auxiliary signal standardizes $u_{ikt}$ over valid tokens of the same response, separately for each component, and averages over the three components:
\[
 S_{it}=\frac{1}{3}\sum_{k=1}^{3}
 \frac{u_{ikt}-\operatorname{mean}_{t'}u_{ikt'}}
 {\sqrt{\operatorname{var}_{t'}u_{ikt'}+10^{-6}}},
 \qquad
 A^{\mathrm{RTT}}_{it}=\operatorname{clip}\!\left(A^{\mathrm{GRPO}}_{it}+0.5S_{it},-3,3\right).
\]
Here, $t'$ ranges over valid response tokens. Normalization is performed within each response rather than across responses; the resulting signal is added to the GRPO advantage.

\subsection{Adapted \texorpdfstring{GD$^2$PO}{GD2PO}}\label{sec:appendix-gd2po}

We extend the response-level conflict filtering in GD$^2$PO~\citep{liu2026gd2po} to the three task rewards. Within each prompt's group $\mathcal G$ of eight responses, we compute
\[
 a_i^k=\frac{r_i^k-\operatorname{mean}_{j\in\mathcal G}r_j^k}
 {\operatorname{std}_{j\in\mathcal G}r_j^k+10^{-4}}.
\]
ProductAttr weights these values by $(0.4,0.4,0.2)$ for weight, dimensions, and foldability; Nutrition5k uses their unweighted sum. The hard-conflict rule masks a response if its valid component advantages include both a value above $10^{-8}$ and a value below $-10^{-8}$, ignoring missing components. Let $m_i$ be 1 for a retained response and 0 otherwise. The response-level policy advantage before token normalization is
\[
 \rho_{\mathcal G}=\frac{1}{8}\sum_{j\in\mathcal G}m_j,
 \qquad
 A_i^{\mathrm{GD^2PO}}=m_i\rho_{\mathcal G}\sum_{k=1}^{3}v_k a_i^k,
\]
where $v=(0.4,0.4,0.2)$ for ProductAttr and $v=(1,1,1)$ for Nutrition5k. We broadcast this value to response tokens, whiten only valid tokens of retained responses, and clip to $[-3,3]$. Masked responses retain zero policy-gradient advantage. A fully filtered group contributes zero policy-gradient advantage; if the entire batch is filtered, whitening is skipped. We do not resample after filtering, so each rollout batch remains 32 prompts with eight responses each. Masking affects the policy-gradient term; the separate KL loss remains active.

\section{Complete Results and Statistical Tests}

\subsection{Training-Seed Replication and Dialogue State Tracking}\label{sec:appendix-replication-sgd}

\begin{table}[H]
\centering\small\color{black}
\caption{ProductAttr summary across three training seeds (\%): means are reported.}
\label{tab:product-training-seeds}
\begin{tabular}{lccc}
\toprule
Method & Weight RMAE $\downarrow$ & Dimensions RMAE $\downarrow$ & Fold F1 $\uparrow$\\
\midrule
GRPO & $28.35$ & $21.21$ & $90.00$\\
GRPO + Ours & $\bestmean{26.10}$ & $\bestmean{20.16}$ & $\bestmean{91.72}$\\
\bottomrule
\end{tabular}
\end{table}

\begin{table}[H]
\centering\small
\caption{SGD test results (\%; higher is better). All methods report the mean and sample standard deviation over decoding seeds~42, 43, and~44 for one fixed model checkpoint per method, not variation across training seeds. Approximately 90\% of Base outputs fail the frozen output-conformance checks and remain in the reported denominators; the Base-to-trained gap therefore includes output-format compliance.}
\label{tab:sgd-full-results}
\begin{tabular}{lcc}
\toprule
Method & Joint Goal Accuracy $\uparrow$ & Average Goal Accuracy $\uparrow$\\
\midrule
Base (No-RL) & $5.77\tablesd{0.63}$ & $7.54\tablesd{0.40}$\\
GRPO & $75.31\tablesd{0.06}$ & $94.48\tablesd{0.02}$\\
+ Ours & $81.31\tablesd{0.11}$ & $96.48\tablesd{0.01}$\\
\bottomrule
\end{tabular}
\end{table}

\subsection{Supplementary ProductAttr Results}\label{sec:appendix-product-results}

\begin{table}[H]
\centering
\small
\caption{Gap comparison on ProductAttr-Eval (\%), using fixed endpoints and three evaluation seeds.}
\label{tab:product-gap}
\begin{tabular}{lcccc}
\toprule
Gap & Weight RMAE $\downarrow$ & Dimensions RMAE $\downarrow$ & Fold F1 $\uparrow$ & Valid-output rate $\uparrow$\\
\midrule
0.50 & $\bestmean{25.12}\tablesd{0.34}$ & $\bestmean{19.68}\tablesd{0.06}$ & $91.31\tablesd{0.42}$ & $99.67\tablesd{0.14}$\\
0.75 & $27.92\tablesd{0.14}$ & $20.82\tablesd{0.06}$ & $\bestmean{93.00}\tablesd{0.48}$ & $\bestmean{100.00}\tablesd{0.00}$\\
1.00 & $28.77\tablesd{0.75}$ & $21.62\tablesd{0.18}$ & $92.14\tablesd{0.37}$ & $\bestmean{100.00}\tablesd{0.00}$\\
\bottomrule
\end{tabular}
\end{table}
Larger gap thresholds increase weight and dimension errors but improve Fold F1 relative to the default threshold of 0.50, which is used in the main experiments and component ablations.

\paragraph{FiveTask extension evaluation.}
To examine credit assignment after adding seasonality and festival attributes, we compare matched GRPO and \method{} on the same ProductAttr-Eval split. Table~\ref{tab:product-fivetask} shows that \method{} reduces weight and dimension RMAE by $3.18$ and $2.28$ percentage points, respectively, and increases sample-level seasonality Dice by $1.80$ percentage points. Fold F1 is approximately unchanged, and festival macro-F1 increases by $0.32$ percentage points. Seasonality Dice matches the task's training reward and measures per-product label-set agreement, including the nonseasonal label.

\begin{table}[H]
\centering
\scriptsize
\caption{ProductAttr FiveTask extension (\%), using a fixed single training checkpoint and three evaluation seeds. S-Dice is sample-level seasonal Dice; F-Macro is Macro-F1 over eight holiday classes.}
\label{tab:product-fivetask}
\resizebox{\textwidth}{!}{%
\begin{tabular}{lrrrrr}
\toprule
Method & Weight RMAE $\downarrow$ & Dimensions RMAE $\downarrow$ & Fold F1 $\uparrow$ & S-Dice $\uparrow$ & F-Macro $\uparrow$\\
\midrule
GRPO & $29.36\tablesd{0.57}$ & $21.98\tablesd{0.38}$ & $\bestmean{91.60}\tablesd{0.06}$ & $89.43\tablesd{0.47}$ & $96.41\tablesd{0.30}$\\
\method{} & $\bestmean{26.18}\tablesd{0.43}$ & $\bestmean{19.70}\tablesd{0.04}$ & $91.51\tablesd{0.32}$ & $\bestmean{91.23}\tablesd{0.23}$ & $\bestmean{96.73}\tablesd{0.80}$\\
\bottomrule
\end{tabular}}
\end{table}

\subsection{Statistical Tests}\label{sec:appendix-statistics}

Table~\ref{tab:product-training-seeds} summarizes replication across three ProductAttr training seeds. The intervals below instead quantify evaluation variability for the fixed models trained with seed 42.

For ProductAttr, we use a hierarchical paired bootstrap over evaluation seeds and products, with \texttt{item\_id+sku\_id} as the pairing unit. Each replicate draws three evaluation seeds with replacement and resamples the 800 products, sharing the sampled product indices across methods, metrics, and selected seeds. We recompute the metrics for each sampled seed and average across seeds. We use $B=20{,}000$ replicates and percentile 95\% confidence intervals. Reported standard deviations across evaluation seeds use ddof=1.

For \method{} minus GRPO, the differences in weight RMAE, dimension RMAE, and Fold F1 are $-5.74$, $-2.80$, and $+1.76$ percentage points, respectively. Their 95\% intervals are $[-8.48,-3.09]$, $[-4.99,-1.69]$, and $[-1.27,4.85]$. The continuous-attribute improvement intervals exclude zero, whereas the Fold interval includes zero. These intervals do not cover training randomness; other between-method differences are descriptive unless an interval is explicitly reported.

\section{Additional Baselines, Ablations, and Sensitivity Analysis}

\subsection{Cross-Family Evaluation}\label{sec:cross-model-results}

Table~\ref{tab:scale-replication} reports the Gemma3-4B-IT comparison. GRPO and GRPO + \method{} use the same initialization, data, training budget, and decoding configuration. Larger Qwen dense and MoE results are reported in Table~\ref{tab:host-compatibility}, panel~b; the MoE model has approximately 35B total parameters and activates approximately 3B per token.

\begin{table}[H]
\centering\small
\caption{Cross-family evaluation on Gemma3-4B-IT using ProductAttr-Eval (\%). Means and sample standard deviations use three evaluation seeds for fixed training checkpoints, following Section~\ref{sec:exp-setup}.}
\label{tab:scale-replication}
\setlength{\tabcolsep}{5pt}
\begin{tabular}{lccc}
\toprule
Method & Weight RMAE $\downarrow$ & Dimensions RMAE $\downarrow$ & Fold F1 $\uparrow$\\
\midrule
GRPO & $\bestmean{51.53}\tablesd{0.74}$ & $133.86\tablesd{37.96}$ & $55.91\tablesd{1.63}$\\
GRPO + Ours & $52.25\tablesd{2.27}$ & $\bestmean{50.88}\tablesd{24.86}$ & $\bestmean{59.15}\tablesd{1.14}$\\
\bottomrule
\end{tabular}
\end{table}

For Gemma3-4B-IT, \method{} improves dimension RMAE and Fold F1, while weight RMAE increases from 51.53 to 52.25. Dimension errors have large standard deviations across decoding seeds; even after removing any single extreme outlier item, dimension RMAE drops by 64.56--92.27 points.

\subsection{Online Routing Granularity}\label{sec:online-routing-controls}

\paragraph{Routing granularity.}
Token-route selects candidates and assigns credit separately at each reasoning token, rather than aggregating attention within a sentence. Both variants use gap threshold 0.50 and the same field-value credit, conflict handling, whitening, training budget, and evaluation protocol. Sentence-level routing has lower weight and dimension errors, while Token-route has slightly higher Fold F1.

\begin{table}[H]
\centering\small
\caption{Online routing-granularity comparison on ProductAttr-Eval (\%). Both variants assign component-specific credit to final values and reasoning; results use three evaluation seeds.}
\label{tab:routing-granularity}
\begin{tabular}{lccc}
\toprule
Variant & Weight RMAE $\downarrow$ & Dimensions RMAE $\downarrow$ & Fold F1 $\uparrow$\\
\midrule
Token-route & $27.01\tablesd{2.01}$ & $20.40\tablesd{0.63}$ & $\bestmean{91.46}\tablesd{0.76}$\\
\method{} (sentence route) & $\bestmean{25.12}\tablesd{0.34}$ & $\bestmean{19.68}\tablesd{0.06}$ & $91.31\tablesd{0.42}$\\
\bottomrule
\end{tabular}
\end{table}

\section{Supplementary Association Analysis}\label{sec:association-appendix}

\subsection{Model-Reference Association Evaluation}\label{sec:model-reference-association}

Table~\ref{tab:model-reference-full} reports the full association metrics. The Base and GPT response sets each contain 1,024 responses to 128 ProductAttr training inputs. Base responses yield 43,183 scored sentences from 961 responses; 63 responses have no locatable reasoning sentences and remain in response-level accounting but do not contribute sentence-level labels. All 1,024 GPT responses yield scored sentences, totaling 7,807. Fifty-five Base responses reach the generation limit; this status is separate from the no-sentence category. The reference model is GPT-5.6 Terra (\texttt{gpt-5.6-terra-2026-07-09}). It labels complete visible reasoning using the fixed component-labeling protocol. Attention and gradient attribution on both sources are obtained by replay through Qwen3.5-4B.

\begin{table}[H]
\centering\small
\caption{Full model-reference association metrics on Base and GPT responses (\%). Precision, recall, and F1 are micro-averaged. Coverage is the percentage of sentences with a nonempty predicted component set. Random matching uses the gated method's per-sentence candidate counts and averages ten random seeds. Top-1 is an ungated attention ablation.}
\label{tab:model-reference-full}
\setlength{\tabcolsep}{4pt}
\begin{tabular}{llrrrr}
\toprule
Source & Association method & Precision & Recall & F1 & Coverage\\
\midrule
Base & Matched random & 34.94 & 33.39 & 34.14 & 88.99\\
 & Keyword heuristic & 90.78 & 46.04 & 61.09 & 48.66\\
 & Gradient gate & 37.29 & 38.85 & 38.05 & 95.91\\
 & Top-1 without gating & 79.59 & 75.41 & 77.44 & 100.00\\
 & \method{} & 78.28 & 74.80 & 76.50 & 88.99\\
\midrule
GPT & Matched random & 43.35 & 27.22 & 33.44 & 79.77\\
 & Keyword heuristic & 94.88 & 48.77 & 64.43 & 65.19\\
 & Gradient gate & 55.01 & 43.75 & 48.74 & 99.83\\
 & Top-1 without gating & 94.25 & 67.31 & 78.53 & 100.00\\
 & \method{} & 95.41 & 59.92 & 73.62 & 79.77\\
\bottomrule
\end{tabular}
\end{table}

We use 2,000 paired bootstrap replicates over 128 product clusters per source. The 95\% F1 intervals for \method{} are [74.66, 78.27] on Base and [72.28, 74.88] on GPT responses. Differences from keyword matching are +15.41 [14.05, 16.80] and +9.19 [7.82, 10.44]. Differences from ungated Top-1 are $-0.94$ [$-1.46$, $-0.22$] and $-4.91$ [$-5.70$, $-4.13$]. Intervals describe variability over sampled products, not uncertainty in reference labels. Each source is evaluated separately.

\paragraph{Primary-component reference protocol.}\label{sec:primary-component-protocol}
The GPT response set contains 7,681 relevant sentences and 126 sentences with empty reference labels. Existing single-component labels are retained; for the 2,557 multi-label sentences, Terra selects one primary component from the original reference set. Uncertain primary labels (909 sentences) remain in scoring. Precision is the number of correct primary-component assignments divided by the total number of emitted assignments across all 7,807 sentences. Assignments to empty-reference sentences count as false positives; abstentions do not enter the precision denominator. \method{} (ungated) selects the highest attention score. \method{} (gated) makes the same selection on accepted sentences and otherwise abstains. Their precision values are 84.28\% and 92.29\%, respectively. Matched random uses the same acceptance mask as the gate and achieves 32.93\% precision. Random controls average ten seeds. This single-primary-label diagnostic is distinct from the multi-label evaluation above.

\subsection{Nutrition5k Prompt Control}

{\color{black}
\paragraph{Nutrition5k prompt control.}
We additionally train GRPO and GRPO + \method{} with an English prompt that removes the original restrictions on the number of checks and revisions, while retaining prediction of total fat, carbohydrates, and protein for the whole dish and the strict three-field JSON output requirement. Both arms use this prompt and evaluate final checkpoint-516 on the same 507 samples with evaluation seeds 42, 43, and 44. Table~\ref{tab:nutrition-prompt-comparison} reports the paired results. This control changes both language and reasoning guidance, so differences cannot be attributed to any one instruction.
\begin{table}[H]
\centering\small\color{black}
\caption{Paired Nutrition5k results with the new prompt: attribute RMAE (\%, lower is better), reported as the mean $\pm$ sample standard deviation across three evaluation seeds.}
\label{tab:nutrition-prompt-comparison}
\begin{tabular}{lccc}
\toprule
Method & Fat & Carbohydrate & Protein\\
\midrule
GRPO & $\bestmean{34.71}\tablesd{0.30}$ & $30.39\tablesd{0.38}$ & $33.01\tablesd{0.31}$\\
GRPO + Ours & $36.29\tablesd{0.75}$ & $\bestmean{29.82}\tablesd{0.68}$ & $\bestmean{31.96}\tablesd{0.03}$\\
\bottomrule
\end{tabular}
\end{table}

}

\clearpage
\section{Sentence-level credit in a complete training response}\label{sec:appendix-case-study}

Figure~3 reproduces every recorded span of one Full rollout in its original order. The example was selected from training step 0, seed 42, to illustrate how credit differs across attributes within one response.

\medskip
\noindent\textbf{Prediction}\quad 50 g; $65 \times 65 \times 35$ mm; non-foldable.\\[2pt]
\noindent\textbf{Reference}\quad 40 g; $125 \times 116 \times 24$ mm; non-foldable.

\medskip
\definecolor{shlow}{HTML}{D6E4EA}
\definecolor{shmid}{HTML}{91B9C9}
\definecolor{shhigh}{HTML}{457D97}
\definecolor{gateborder}{HTML}{2A5B72}
\definecolor{advposbg}{HTML}{F1F6F8}
\definecolor{advnegbg}{HTML}{FDF2F0}
\definecolor{advpostxt}{HTML}{2B5B70}
\definecolor{advnegtxt}{HTML}{B03A2E}
\definecolor{advgray}{HTML}{6E737B}

\newcommand{\shbox}[2]{\colorbox{#1}{\parbox[c][11pt][c]{24pt}{\centering\footnotesize #2}}}
\newcommand{\shgate}[3]{\setlength{\fboxsep}{0.6pt}\setlength{\fboxrule}{1pt}\fcolorbox{gateborder}{#1}{\parbox[c][11pt][c]{24pt}{\centering\footnotesize\textcolor{#2}{#3}}}}
\newcommand{\advgraybox}[1]{\parbox[c][11pt][c]{32pt}{\centering\footnotesize\textcolor{advgray}{#1}}}
\newcommand{\advpos}[1]{\colorbox{advposbg}{\parbox[c][11pt][c]{32pt}{\centering\footnotesize\textcolor{advpostxt}{#1}}}}
\newcommand{\advneg}[1]{\colorbox{advnegbg}{\parbox[c][11pt][c]{32pt}{\centering\footnotesize\textcolor{advnegtxt}{#1}}}}

\begingroup\small
\renewcommand{\arraystretch}{1.35}
\setlength{\tabcolsep}{3pt}
\noindent\begin{tabularx}{\linewidth}{@{}l X ccc ccc@{}}
\toprule
\textbf{Span} & \textbf{Complete recorded text} & \multicolumn{3}{c}{\textbf{Anchor share (\%)}} & \multicolumn{3}{c}{\textbf{Final advantage}} \\
 & & \textbf{Weight} & \textbf{Dim.} & \textbf{Fold} & \textbf{GRPO} & \textbf{GDPO} & \textbf{Full} \\
\midrule
\textbf{S00} & Analysis of the product image and listing data reveals that these are compact, bone-conduction style TWS earbuds in a purple, glowing charging case. & \shbox{shlow}{29.9} & \shbox{shmid}{39.2} & \shbox{shlow}{30.9} & \advgraybox{+1.623} & \advgraybox{+0.715} & \advpos{+1.645} \\
\midrule
\textbf{S01} & The listing specifies a quantity, though the SKU description says ``Earbuds'' (plural), the attributes list ``Number Of Drivers: 2'' implying a single unit set is the standard packaging. & \shbox{shlow}{33.8} & \shbox{shmid}{47.8} & \shbox{shlow}{18.4} & \advgraybox{+1.623} & \advgraybox{+0.715} & \advpos{+1.645} \\
\midrule
\textbf{S02} & Standard TWS cases are usually 1-2 units per package, but without a specific quantity multiplier in the SKU details, I will estimate based on a single unit set (1 pair of earbuds) which is the minimum viable shipment. & \shbox{shmid}{37.3} & \shbox{shmid}{41.0} & \shbox{shlow}{21.7} & \advgraybox{+1.623} & \advgraybox{+0.715} & \advpos{+1.645} \\
\midrule
\textbf{S03} & The case is described as having ``Size: 60'' (likely diameter or width) and is visibly rounded and smooth. & \shbox{shlow}{21.0} & \shgate{shhigh}{white}{71.8} & \shbox{shlow!50}{7.2} & \advgraybox{+1.623} & \advgraybox{+0.715} & \advneg{$-$0.301} \\
\midrule
\textbf{S04} & Based on the 60mm measurement and the plastic material, the case is approximately 60mm wide and 60mm deep, but slightly taller (height) when closed, likely around 50-55mm. & \shbox{shlow}{22.5} & \shgate{shhigh}{white}{72.8} & \shbox{shlow!50}{4.7} & \advgraybox{+1.623} & \advgraybox{+0.715} & \advneg{$-$0.301} \\
\midrule
\textbf{S05} & The packaging will likely be a simple blister pack or a small cardboard box (polybag) to protect the hard plastic case. & \shbox{shlow}{27.4} & \shgate{shmid}{black}{55.1} & \shbox{shlow}{17.6} & \advgraybox{+1.623} & \advgraybox{+0.715} & \advneg{$-$0.301} \\
\bottomrule
\end{tabularx}
\endgroup

\medskip
\noindent\small Figure 3. Complete recorded text, anchor-attention shares and final advantages. Attention shares are normalized within each row. Outlines indicate components retained by the gate. Advantages are shown to three decimals. The response continues on the next page.

\smallskip
\noindent\small Full values are logged after final normalization and clipping. GRPO is the captured uniform baseline. GDPO is recomputed on the same complete batch of 256 responses with reward weights 0.4/0.4/0.2 for weight, dimensions and foldability.
\normalsize

\clearpage
\subsection*{F\quad Sentence-level credit (continued)}

\begingroup\small
\renewcommand{\arraystretch}{1.28}
\setlength{\tabcolsep}{3pt}
\noindent\begin{tabularx}{\linewidth}{@{}l X ccc ccc@{}}
\toprule
\textbf{Span} & \textbf{Complete recorded text} & \multicolumn{3}{c}{\textbf{Anchor share (\%)}} & \multicolumn{3}{c}{\textbf{Final advantage}} \\
 & & \textbf{Weight} & \textbf{Dim.} & \textbf{Fold} & \textbf{GRPO} & \textbf{GDPO} & \textbf{Full} \\
\midrule
\textbf{S06} & A single unit of this size is quite small and lightweight, typically weighing between 30g and 60g depending on the case materials and included silicone tips/charger cable. & \shgate{shhigh}{white}{80.6} & \shbox{shlow!70}{14.2} & \shbox{shlow!40}{5.3} & \advgraybox{+1.623} & \advgraybox{+0.715} & \advpos{+1.681} \\
\midrule
\textbf{S07} & Regarding dimensions, the bounding box would encompass the case and its immediate protective packaging. & \shbox{shlow}{20.7} & \shgate{shhigh}{white}{62.1} & \shbox{shlow}{17.3} & \advgraybox{+1.623} & \advgraybox{+0.715} & \advneg{$-$0.301} \\
\midrule
\textbf{S08} & A standard 1-unit box for these earbuds is roughly 65mm x 65mm x 35mm. & \shbox{shlow!70}{13.2} & \shgate{shhigh}{white}{84.0} & \shbox{shlow!30}{2.8} & \advgraybox{+1.623} & \advgraybox{+0.715} & \advneg{$-$0.301} \\
\midrule
\textbf{S09} & Ordering by the criteria $L\ge W\ge H$, this translates to 65mm length, 65mm width, and 35mm height. & \shbox{shlow!60}{10.3} & \shgate{shhigh}{white}{84.2} & \shbox{shlow!40}{5.5} & \advgraybox{+1.623} & \advgraybox{+0.715} & \advneg{$-$0.301} \\
\midrule
\textbf{S10} & The weight is estimated at roughly 50 grams for a single unit with standard protective packaging. & \shgate{shhigh}{white}{85.3} & \shbox{shlow!60}{10.4} & \shbox{shlow!40}{4.3} & \advgraybox{+1.623} & \advgraybox{+0.715} & \advpos{+1.681} \\
\midrule
\textbf{S11} & The items are rigid plastic with no hinges, straps, or flexible components that can be compacted. & \shbox{shlow!50}{8.9} & \shbox{shlow}{22.1} & \shgate{shhigh}{white}{68.9} & \advgraybox{+1.623} & \advgraybox{+0.715} & \advpos{+0.720} \\
\midrule
\textbf{S12} & The case must remain closed to prevent damage and to hold the electronics. & \shbox{shlow!60}{10.0} & \shbox{shlow}{30.5} & \shgate{shmid}{black}{59.4} & \advgraybox{+1.623} & \advgraybox{+0.715} & \advpos{+0.720} \\
\midrule
\textbf{S13} & Therefore, the item cannot be folded or rolled. & \shbox{shlow!30}{3.9} & \shbox{shlow!60}{10.7} & \shgate{shhigh}{white}{85.4} & \advgraybox{+1.623} & \advgraybox{+0.715} & \advpos{+0.720} \\
\midrule
\textbf{S14} & \texttt{\`{}\`{}\`{}json} & \shbox{shlow}{17.2} & \shbox{shlow!60}{11.2} & \shgate{shhigh}{white}{71.6} & \advgraybox{+1.623} & \advgraybox{+0.715} & \advpos{+0.720} \\
\bottomrule
\end{tabularx}
\endgroup

\medskip
\noindent\small Figure 3 (continued). S06--S13 complete the reasoning. S14 is the recorded JSON delimiter. Attention percentages may sum to 99.9\% or 100.1\% after rounding. GRPO and GDPO allocate one response-level value to every span.

\medskip
\noindent\begin{tabularx}{\linewidth}{@{}p{0.46\linewidth} X@{}}
\textbf{Generated JSON}\par\smallskip
{\footnotesize\ttfamily
\{\par
\ \ "item\_weight\_cn": 50.0,\par
\ \ "item\_length\_cn": 65.0,\par
\ \ "item\_width\_cn": 65.0,\par
\ \ "item\_height\_cn": 35.0,\par
\ \ "final\_fold\_flag": 0\par
\}}
&
\vspace{8pt}\small GDPO assigns $+0.715$ throughout. Full assigns $-0.301$ to the dimensions spans, $+1.681$ to the weight spans and $+0.720$ to the foldability spans. S00--S02 use the response-level fallback before Full's final normalization.
\end{tabularx}
\normalsize

\clearpage

\section{Additional experiments and measurement protocols}\label{sec:appendix-g}

\subsection*{G.1 Attention-weighted credit-combination baseline and pre-routing whitening analysis}

We report the new matched ProductAttr runs separately from the historical runs in the submitted paper. Each new run uses Qwen3.5-4B, training seed 42, and 600 updates. The two credit-combination runs share the initial model and training hyperparameters. Each checkpoint is evaluated on 800 products with decoding seeds 42, 43, and 44. A reported metric is the mean of three separately computed corpus-level scores; the three decodings are not independent training replications.

\textbf{Table 18. Attention-weighted credit-combination baseline.} RMAE and F1 are percentages. All 4,800 predictions across these two checkpoints are valid. WA averages all component advantages using anchor-attention weights without gating. Full uses the gate and the sign-aware combination rule.

\begingroup\small
\renewcommand{\arraystretch}{1.15}
\noindent\begin{tabularx}{\linewidth}{@{}>{\hsize=1.19192\hsize\raggedright\arraybackslash}X>{\hsize=0.90909\hsize\raggedright\arraybackslash}X>{\hsize=1.03030\hsize\raggedright\arraybackslash}X>{\hsize=0.86869\hsize\raggedright\arraybackslash}X@{}}
\toprule
Method & Weight RMAE $\downarrow$ & Dimension RMAE $\downarrow$ & Fold F1 $\uparrow$ \\
\midrule
Full & 25.76 & 19.56 & 92.60 \\
Weighted-All (WA) & 26.91 & 19.95 & 92.54 \\
\bottomrule
\end{tabularx}\endgroup
\par\medskip

Full minus WA is -1.15 points [-3.97, 1.94] in weight RMAE, -0.39 [-0.93, 0.02] in dimension RMAE, and +0.06 [-1.98, 2.10] in Fold F1. Intervals use 20,000 paired bootstrap samples over products and the three observed decoding seeds, conditional on the fixed checkpoints. All three intervals include zero. In first-step routing statistics on each run's generated responses, WA routes every sentence and averages positive and negative component advantages together for 67.63\% of sentences, whereas Full routes a sentence only when its attention association passes the gate (with 10.10\% of sentences falling back to the response-level advantage and 5.25\% having opposite-sign gated candidates, for which Full selects the most strongly associated component).

\textbf{Design motivation and offline replay of pre-routing whitening (Equation 2).} Following GDPO (Liu et al., 2026b), Equation 1 first normalizes each reward component within each prompt group to decouple heterogeneous reward scales. However, whereas GDPO immediately aggregates component advantages into a single response-level scalar and applies batch whitening only to that sum, AnchorCredit routes distinct component advantages to different sentences, averages multiple same-sign candidates (Equation 7), compares candidate signs for conflict handling, and falls back to the response-level advantage on unrouted sentences. Even after group normalization, different components can exhibit unequal effective batch variances and slight mean shifts: for example, binary attributes such as foldability produce more zero-variance prompt groups (when all rollouts in a group predict the same binary label, yielding zero group-relative advantage) than continuous attributes (weight and dimensions). Pre-routing whitening (Equation 2) calibrates each component-advantage column and the fallback advantage to a common zero-mean, unit-variance batch scale before routing, same-sign averaging, and sign comparisons. In an offline replay of 8,992 reasoning sentences holding responses, attention, and raw advantages fixed, enabling pre-routing whitening increases mixed-sign cases from 278 to 469, reverses the credit sign for 93 sentences, and changes credit from zero to nonzero for 3,072 sentences.

\clearpage
\subsection*{G.2 Attention-layer aggregation against existing GPT labels}

We freeze Qwen3.5-4B and replay 1,024 GPT-generated responses from 128 products, containing 7,807 reasoning sentences. Only the set of full-attention layers used for aggregation changes. The eligible zero-based layer indices are 3, 7, 11, 15, 19, 23, 27, and 31. Linear-attention layers do not expose the explicit attention matrices required by this analysis and are excluded.

The existing reference annotations were produced by GPT-5.6 Terra (gpt-5.6-terra-2026-07-09). A sentence can be associated with more than one component. We predict the component with the highest attention after averaging over the specified layers and count a hit when that component belongs to the reference set. All 7,807 sentences are evaluated, including 126 empty-reference sentences, which count as misses. The remaining references comprise 5,124 singleton and 2,557 multilabel sets. Uncertainty flags are retained rather than used to filter sentences. No new external-model annotations are used.

\textbf{Table 19(a). Qwen Top-1 agreement under different layer aggregations.} All rows use the same responses, model, reference labels, and denominator. This diagnostic uses a prediction for every sentence; it does not measure selective prediction after gating.

\begingroup\small
\renewcommand{\arraystretch}{1.15}
\noindent\begin{tabularx}{\linewidth}{@{}>{\hsize=0.45455\hsize\raggedright\arraybackslash}X>{\hsize=1.46465\hsize\raggedright\arraybackslash}X>{\hsize=1.11111\hsize\raggedright\arraybackslash}X>{\hsize=0.96970\hsize\raggedright\arraybackslash}X@{}}
\toprule
Layer group & Zero-based indices & Hits / sentences & Agreement \\
\midrule
All & 3, 7, 11, 15, 19, 23, 27, 31 & 7,359 / 7,807 & 94.26\% \\
Early & 3, 7, 11 & 7,265 / 7,807 & 93.06\% \\
Middle & 15, 19, 23 & 7,104 / 7,807 & 91.00\% \\
Late & 27, 31 & 6,557 / 7,807 & 83.99\% \\
\bottomrule
\end{tabularx}\endgroup
\par\medskip

All-layer averaging exceeds Early, Middle, and Late by 1.20, 3.27, and 10.27 percentage points. Corresponding 95\% paired bootstrap intervals are [0.77, 1.66], [2.45, 4.13], and [9.53, 11.05]. We resample the 128 products jointly across methods for 2,000 replicates (random seed 20260921), keeping all responses and sentences of a sampled product together.

The result supports all-layer averaging as a default for this sentence--component ranking diagnostic. It does not establish an optimal layer aggregation for every metric, model, or downstream training objective. In particular, membership in a multilabel reference set differs from the single dominant-component target used in Table 6 of the submitted paper. The percentages in those two evaluations should not be compared as if they used the same target.

The supplied annotation IDs and input hashes were matched to the frozen response records before scoring. All four rows are recomputed within one replay. The historical all-layer result has 7,358 hits; its predictions agree with this replay on 7,802/7,807 sentences. We use the within-replay results throughout Table 19(a) rather than mixing historical and recomputed predictions.

\clearpage

\subsection*{G.2 Attention-layer aggregation (continued)}

\paragraph{Frozen-response diagnostic on a hybrid Gemma model.}
We replay the same 1,024 GPT responses and 7,807 reasoning sentences through frozen Gemma4 E4B (\texttt{google/gemma-4-E4B-it}). The existing GPT reference labels and answer texts are unchanged. We use Gemma's native image processor, chat template, and tokenizer, remapping sentence and output-value spans by character offsets rather than reusing Qwen token positions. The model has 42 text-attention layers: 35 sliding-window layers (window size 512) and seven global layers. Native masks and key/value sharing are preserved. One forward per response supplies all layer groups; no answers are regenerated or newly labeled.

\textbf{Table 19(b). Gemma layer aggregation on the same frozen responses.} All-explicit includes local and global layers; Global-only uses the seven global layers, consistent with the global-layer convention in Section~4.2. Top-1 agreement uses the same multilabel membership target and all 7,807 sentences as panel (a), without gating.

\begingroup\small
\renewcommand{\arraystretch}{1.12}
\noindent\begin{tabularx}{\linewidth}{@{}l>{\raggedright\arraybackslash}Xrr@{}}
\toprule
Layer group & Zero-based indices & Layers & Agreement \\
\midrule
All-explicit & 0--41 & 42 & 93.77\% \\
Early-explicit & 0--13 & 14 & 75.09\% \\
Middle-explicit & 14--27 & 14 & 92.94\% \\
Late-explicit & 28--41 & 14 & 83.99\% \\
\midrule
Global-only & 5, 11, 17, 23, 29, 35, 41 & 7 & 93.39\% \\
Early-global & 5, 11 & 2 & 81.54\% \\
Middle-global & 17, 23, 29 & 3 & 91.79\% \\
Late-global & 35, 41 & 2 & 87.69\% \\
\bottomrule
\end{tabularx}\endgroup
\par\medskip

All-explicit exceeds the best of its three depth subsets, Middle-explicit, by 0.83 percentage points (95\% paired product-bootstrap interval [0.19, 1.43]). Its difference from Global-only is 0.38 points [-0.22, 1.00]; this interval establishes neither a clear advantage nor equivalence. Within the global layers, averaging all seven exceeds Early-global, Middle-global, and Late-global by 11.85 [11.06, 12.64], 1.60 [0.87, 2.28], and 5.70 [4.56, 6.80] points. The global depth subsets differ in both depth and layer count. All intervals use 2,000 paired product-cluster bootstrap replicates with seed 20260921 and are unadjusted across comparisons.

These results support aggregation across depth on this non-Qwen hybrid architecture, without establishing that including local layers is universally preferable. They measure association ranking on fixed GPT responses, not the quality of Gemma-generated answers, the effectiveness of gating, or downstream gains after Gemma RL training. The Gemma3 endpoint results in Table~14 are a separate experiment.

\paragraph{Verification.}
All 42 layers were captured for every response. Each GPU's real-input validation matched sparse anchor-attention rows to native eager attention with maximum absolute difference zero. The 32-response pilot preceded the full run; all 1,024 responses and 7,807 sentences were retained. Recomputing all eight group means from stored per-layer scores and rescoring locally reproduced the cloud results. There were no all-zero predictions or positive-score ties. The checkpoint revision is \texttt{4ecc2ba63bcac23ba4a767fc87025b1450f84c9c}.

\clearpage

\subsection*{G.3 Computational cost}

\subsubsection*{Direct measurement of the additional attention forward}

To measure the cost without comparing outputs of different lengths from different policies, we time the no-gradient attention-extraction forward within AnchorCredit updates on fixed responses. For each of three input-length ranges, three disjoint packets (four prompts, eight responses per prompt) each undergo 10 warm-up updates followed by 50 measured updates on eight A100 GPUs. Model initialization is reset between runs. The Full arm therefore contributes 90 warm-up and 450 measured updates. All results below exclude warm-up.

\textbf{Table 20. Extra attention-forward time within Full updates.} Seconds are means over 150 measured updates per length range. The pooled percentage is the sum of forward times divided by the sum of update times.

\begingroup\small
\renewcommand{\arraystretch}{1.15}
\noindent\begin{tabularx}{\linewidth}{@{}>{\hsize=0.88889\hsize\raggedright\arraybackslash}X>{\hsize=0.98990\hsize\raggedright\arraybackslash}X>{\hsize=1.03030\hsize\raggedright\arraybackslash}X>{\hsize=1.09091\hsize\raggedright\arraybackslash}X@{}}
\toprule
Input length & Extra forward (s) & Full update (s) & Forward / update \\
\midrule
Short & 1.748 & 14.349 & 12.18\% \\
Medium & 2.167 & 16.690 & 12.98\% \\
Long & 2.728 & 21.330 & 12.79\% \\
Pooled, 450 updates & --- & --- & 12.69\% \\
\bottomrule
\end{tabularx}\endgroup
\par\medskip

The numerator uses the maximum measured model-forward duration across participating workers. The denominator starts at advantage computation and ends after actor updating and synchronization. Generation, verifier rewards, and reference-model inference are outside this interval. Thus 12.69\% is a component's share of Full update time, not the percentage increase of the complete pipeline over GRPO. Routing, communication, and other attribution work are not part of the model-forward numerator.

For a separate end-to-end observation over the actual 600-step Full trajectory, the entire attribution prepass, including forward, routing, communication, and driver finalization, totals 12,798.27 s out of 85,306.61 s of recorded step wall time (15.00\%). In-step checkpoint work is included; initialization and endpoint evaluation are excluded. This broader measurement has a different numerator and denominator from Table 20.

\subsubsection*{Historical training-job measurements}

\textbf{Table 21. Observed training cost on Qwen3.5-4B.} Each method requests eight A100 GPUs. Steady-window timing uses 370 comparable recorded steps, while complete job wall time sums the recorded running-to-terminal intervals across the training segments. Each policy generates its own responses in these runs.

\begingroup\small
\renewcommand{\arraystretch}{1.15}
\noindent\begin{tabularx}{\linewidth}{@{}>{\hsize=1.74242\hsize\raggedright\arraybackslash}X>{\hsize=0.62879\hsize\raggedright\arraybackslash}X>{\hsize=0.62879\hsize\raggedright\arraybackslash}X@{}}
\toprule
Measurement & GRPO & AnchorCredit \\
\midrule
Steady-window time per step & 106.7 s & 144.1 s \\
Complete training-job wall time & 22.71 h & 27.16 h \\
Nominal A100 GPU-hours & 181.7 & 217.3 \\
Peak GPU memory / training process & 37.64 GiB & 37.73 GiB \\
\bottomrule
\end{tabularx}\endgroup
\par\medskip

Nominal GPU-hours equal eight times job wall time. Device binding/release times were not recorded, so these are not measurements of actual allocated GPU-hours. The historical +35\% steady-step difference and +20\% job-clock difference include output-length and execution effects and should not be attributed entirely to the additional forward. AnchorCredit requires no additional network, trainable parameters, or backward pass.

\clearpage

\subsection*{G.3 Computational cost (continued)}

\subsubsection*{Checkpoints selected under a common job-clock budget}

The common budget is the smaller of the two historical endpoint job clocks: 22.648 h from recorded running status to checkpoint registration, accumulating preceding training segments. At each budget threshold, we select the latest registered checkpoint available, without consulting its evaluation performance. Checkpoints are saved every 100 steps and at the endpoint. The two clocks select GRPO step 599 and AnchorCredit step 400 at the full budget.

\textbf{Table 22. Performance within the common 22.65 h job-clock budget.} Metrics are means over the same 800 products and decoding seeds 42, 43, and 44. The budget is an upper bound; the selected checkpoints need not consume equal time.

\begingroup\small
\renewcommand{\arraystretch}{1.15}
\noindent\begin{tabularx}{\linewidth}{@{}>{\hsize=1.48485\hsize\raggedright\arraybackslash}X>{\hsize=0.60101\hsize\raggedright\arraybackslash}X>{\hsize=0.84848\hsize\raggedright\arraybackslash}X>{\hsize=0.95455\hsize\raggedright\arraybackslash}X>{\hsize=1.14899\hsize\raggedright\arraybackslash}X>{\hsize=0.98990\hsize\raggedright\arraybackslash}X>{\hsize=0.97222\hsize\raggedright\arraybackslash}X@{}}
\toprule
Method & Step & Clock (h) & Weight $\downarrow$ & Dimension $\downarrow$ & Fold F1 $\uparrow$ & Validity $\uparrow$ \\
\midrule
GRPO & 599 & 22.65 & 30.86 & 22.48 & 89.55 & 99.88 \\
AnchorCredit & 400 & 19.15 & 26.13 & 20.59 & 90.07 & 99.38 \\
\bottomrule
\end{tabularx}\endgroup
\par\medskip

Weight and Dimension denote RMAE (\%). AnchorCredit minus GRPO, computed before rounding, is -4.74 points [-7.68, -1.96] for weight RMAE, -1.88 [-4.00, -0.71] for dimension RMAE, +0.52 [-2.27, 3.28] for Fold F1, and -0.50 [-1.08, 0.04] for validity. Intervals use 20,000 paired product/decoding-seed bootstrap replicates conditional on these two training trajectories. This is a descriptive historical job-clock comparison; it is not an exact allocation-matched GPU-hour experiment.

\subsubsection*{Inference measurements}

The trained model architecture is unchanged. Inference uses ordinary generation, with no additional attribution forward, gating, or reward computation. We measured both trained policies on the same L20 GPU at concurrency 16, using 200 products per round and three decoding seeds. Table 23 averages the corresponding measurements over the three rounds.

\textbf{Table 23. Inference latency and output lengths.} Mean request latency and complete-batch time measure different aspects of the output-length distribution.

\begingroup\small
\renewcommand{\arraystretch}{1.15}
\noindent\begin{tabularx}{\linewidth}{@{}>{\hsize=1.74242\hsize\raggedright\arraybackslash}X>{\hsize=0.62879\hsize\raggedright\arraybackslash}X>{\hsize=0.62879\hsize\raggedright\arraybackslash}X@{}}
\toprule
Measurement & GRPO & AnchorCredit \\
\midrule
Mean request latency & 9.39 s & 6.14 s \\
Mean output length & 442 tokens & 263 tokens \\
Mean time to first token & 0.40 s & 0.40 s \\
Complete batch of 200 requests & 135.6 s & 173.6 s \\
\bottomrule
\end{tabularx}\endgroup
\par\medskip

AnchorCredit's average output is shorter and its mean request latency is lower. However, a few requests in each round reach the 8,192-token generation limit, making its complete-batch time longer. These observations do not support a general inference-speedup claim. They distinguish an unchanged inference computation graph from runtime differences caused by learned output behavior.

\clearpage

\subsection*{G.4 Public benchmarks and the number of reward components}

\subsubsection*{RAMS event argument extraction}

RAMS contains news documents with event argument annotations (Ebner et al., 2020). Each example requests 1--5 event roles, each scored as a reward component; 57.29\% of test documents request more than three roles. We rescore the existing checkpoint-599 outputs from GRPO and AnchorCredit on all 871 test documents and 2,023 gold links, with decoding seeds 42, 43, and 44. This adds a public-benchmark evaluation in a distinct domain without representing the rescoring as a new training run.

\textbf{Table 24. RAMS results.} All scores are percentages; differences are AnchorCredit minus GRPO in percentage points. Corpus-level Link F1 is verified against the official scorer. Per-role span F1 is the mean of within-document role-level multiset span F1; joint accuracy requires all requested role arguments in a document to be correct. Format validity measures compliance with the output schema and span constraints.

\begingroup\small
\renewcommand{\arraystretch}{1.15}
\noindent\begin{tabularx}{\linewidth}{@{}>{\hsize=1.19192\hsize\raggedright\arraybackslash}X>{\hsize=0.58586\hsize\raggedright\arraybackslash}X>{\hsize=0.74747\hsize\raggedright\arraybackslash}X>{\hsize=1.47475\hsize\raggedright\arraybackslash}X@{}}
\toprule
Metric & GRPO & AnchorCredit & Difference [95\% CI] \\
\midrule
Link F1 & 43.87 & 44.31 & +0.45 [-1.23, 2.12] \\
Per-role span F1 & 58.35 & 58.82 & +0.47 [-0.69, 1.61] \\
Joint accuracy & 16.30 & 17.95 & +1.65 [-0.15, 3.52] \\
Format validity & 99.85 & 99.46 & -0.38 [-0.92, 0.11] \\
\bottomrule
\end{tabularx}\endgroup
\par\medskip

Invalid JSON, schema, or spans contribute zero true-positive/predicted links while retaining all gold links. No valid-only filtering is used. Scores are first computed per decoding seed and then averaged. Intervals use 20,000 jointly paired document and observed-seed bootstrap samples, conditional on the two checkpoints; no multiple-metric correction is applied. All four intervals include zero. The positive extraction-score estimates therefore do not establish a statistically significant improvement. Twenty-three documents have no gold links; the corpus Link F1 denominator is nonzero in the observed data and all bootstrap replicates.

\subsubsection*{SGD results by the number of requested slots}

The complete SGD test set has 46,116 frames from 4,201 dialogues. Requested schema size K ranges from 2 to 13 (mean 8.36); each slot is a separate reward component. The $K>3$ stratum contains 44,874 frames (97.31\%), and K$\leq$3 contains 1,242. K is the number of requested fields, not the number of nonempty gold fields or nonzero advantages.

\textbf{Table 25. SGD JGA and its improvement by K.} Gains are AnchorCredit minus GRPO, in percentage points. Both listed decoding seeds have complete coverage of the test frames.

\begingroup\small
\renewcommand{\arraystretch}{1.15}
\noindent\begin{tabularx}{\linewidth}{@{}>{\hsize=0.77273\hsize\raggedright\arraybackslash}X>{\hsize=0.95455\hsize\raggedright\arraybackslash}X>{\hsize=1.30303\hsize\raggedright\arraybackslash}X>{\hsize=0.93939\hsize\raggedright\arraybackslash}X>{\hsize=1.01515\hsize\raggedright\arraybackslash}X>{\hsize=1.01515\hsize\raggedright\arraybackslash}X@{}}
\toprule
Seed & GRPO JGA & AnchorCredit JGA & Overall gain & Gain, $K>3$ & Gain, K$\leq$3 \\
\midrule
42 & 75.32 & 81.21 & +5.89 & +6.00 & +2.11 \\
43 & 75.26 & 81.30 & +6.04 & +6.13 & +2.59 \\
\bottomrule
\end{tabularx}\endgroup
\par\medskip

The overall-gain intervals are [5.39, 6.39] and [5.55, 6.53], from 20,000 paired whole-dialogue bootstrap samples. The difference between high-K and low-K gains is +3.88 [1.13, 6.59] and +3.54 [0.93, 6.14], jointly resampling dialogues across strata. These are natural strata, not randomized choices of K. They establish that the evaluated workload already includes more than three reward components, but do not identify a causal effect of increasing K or measure asymptotic scaling cost.

Reference: Seth Ebner, Patrick Xia, Ryan Culkin, Kyle Rawlins, and Benjamin Van Durme. Multi-Sentence Argument Linking. Proceedings of ACL 2020, pp. 8057--8077.

\end{document}